\documentclass[final,5p,nopreprintline,twocolumn]{elsarticle}
\usepackage{booktabs}
\usepackage[english]{babel}
\usepackage{times}
\usepackage{amsmath}
\usepackage{amssymb}
\usepackage{natbib}
\usepackage{hyperref}
\usepackage{bm}
\usepackage{wrapfig}
\usepackage{xcolor}
\usepackage{mathrsfs}
\usepackage{multirow}
\usepackage{makecell}
\usepackage{arydshln}
\usepackage{graphicx}
\usepackage{subcaption}
\usepackage{algpseudocode}
\usepackage[ruled,lined,linesnumbered]{algorithm2e}

\def\argmax{\mathop{\rm argmax}}%

\begin{document}

\title{Neural Machine Translation: A Review of Methods, Resources, and Tools}

\author[addr1,addr3,addr4]{Zhixing Tan}

\author[addr1,addr3,addr4]{Shuo Wang}

\author[addr1,addr3,addr4]{Zonghan Yang}

\author[addr1,addr3,addr4]{Gang Chen}

\author[addr1,addr3,addr4]{Xuancheng Huang}

\author[addr1,addr3,addr4,addr5]{\\Maosong Sun}

\author[addr1,addr2,addr3,addr4,addr5]{Yang Liu\corref{mycorrespondingauthor}}

\address[addr1]{Department of Computer Science and Technology, Tsinghua University}
\address[addr2]{Institute for AI Industry Research, Tsinghua University}
\address[addr3]{Institute for Artificial Intelligence, Tsinghua University}
\address[addr4]{Beijing National Research Center for Information Science and Technology}
\address[addr5]{Beijing Academy of Artificial Intelligence}

\cortext[mycorrespondingauthor]{Corresponding author.} 
\ead{liuyang2011@tsinghua.edu.cn}

\begin{frontmatter}

\begin{abstract}
Machine translation (MT) is an important sub-field of natural language processing that aims to translate natural languages using computers. In recent years, end-to-end neural machine translation (NMT) has achieved great success and has become the new mainstream method in practical MT systems. In this article, we first provide a broad review of the methods for NMT and focus on methods relating to architectures, decoding, and data augmentation. Then we summarize the resources and tools that are useful for researchers. Finally, we conclude with a discussion of possible future research directions.
\end{abstract}


\end{frontmatter}

\section{Introduction}

\begin{figure*}[!t]
  \centering
  \includegraphics[width=0.8\linewidth]{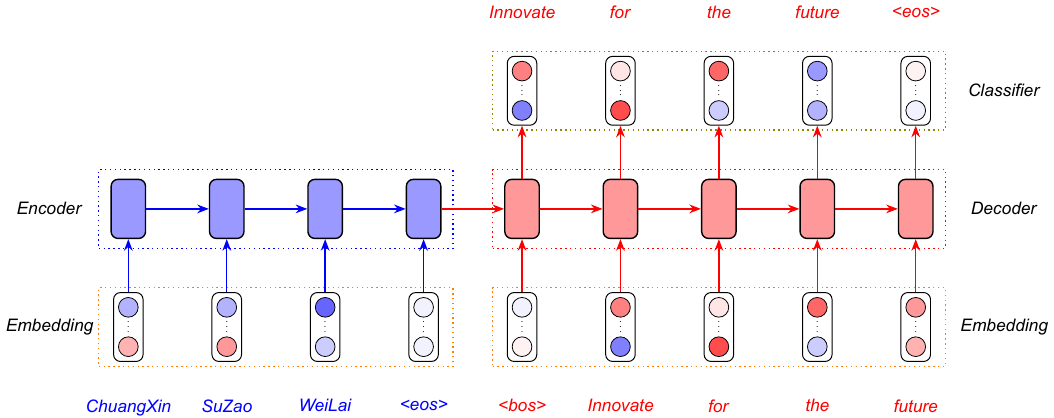}
   \caption{An overview of the NMT architecture, which consists of embedding layers, a classification layer, an encoder network, and a decoder network. We use different colors to distinguish different languages.}
   \label{fig:nmt}
 \end{figure*}

Machine Translation (MT) is an important task that aims to translate natural language sentences using computers. The early approach to machine translation relies heavily on hand-crafted translation rules and linguistic knowledge. As natural languages are inherently complex, it is difficult to cover all language irregularities with manual translation rules. With the availability of large-scale parallel corpora, data-driven approaches that learn linguistic information from data have gained increasing attention. Unlike rule-based machine translation, Statistical Machine Translation (SMT)~\cite{brown1990statistical,koehn2003statistical} learns latent structures such as word alignments or phrases directly from parallel corpora. Incapable of modeling long-distance dependencies between words, the translation quality of SMT is far from satisfactory. With the breakthrough of deep learning, Neural Machine Translation (NMT)~\cite{kal2013rctm,cho2014encdec,sutskever2014seq2seq,bahdanau2015nmt} has emerged as a new paradigm and quickly replaced SMT as the mainstream approach to MT.

Neural machine translation is a radical departure from previous machine translation approaches. On the one hand, NMT employs continuous representations instead of discrete symbolic representations in SMT. On the other hand, NMT uses a single large neural network to model the entire translation process, freeing the need for excessive feature engineering.  The training of NMT is end-to-end as opposed to separately tuned components in SMT. Besides its simplicity, NMT has achieved state-of-the-art performance on various language pairs~\cite{junczys2016neural}. In practice, NMT also becomes the key technology behind many commercial MT systems~\cite{wu2016gnmt,hassan2018achieving}.

As neural machine translation attracts much research interest and grows into an area with many research directions, we believe it is necessary to conduct a comprehensive review of NMT. In this work, we will give an overview of the key ideas and innovations behind NMT. We also summarize the resources and tools that are useful and easily accessible. We hope that by tracing the origins and evolution of NMT, we can stand on the shoulder of past studies, and gain insights into the future of NMT.

The remainder of this article is organized as follows: Section~\ref{sec:methods} will review the methods of NMT. We first introduce the basics of NMT, and then we selectively describe the recent progress of NMT. We focus on methods related to architectures, decoding, and data augmentation. Section~\ref{sec:resources} will summarize the resources such as parallel or monolingual corpora that are publicly available to researchers. Section~\ref{sec:tools} will describe tools that are useful for training and evaluating NMT models. Finally, we conclude and discuss future directions in Section~\ref{sec:conclusion}.

\section{Methods} \label{sec:methods}

As a data-driven approach to machine translation, NMT also embraces the probabilistic framework. Mathematically speaking, the goal of NMT is to estimate an unknown conditional distribution $P(\bm{y}|\bm{x})$ given the dataset $\mathcal{D}$, where $\bm{x}$ and $\bm{y}$ are random variables representing source input and target output, respectively. We strive to answer the three basic questions of NMT:
\begin{itemize}
\item \textit{Modeling.} How to design neural networks to model the conditional distribution?
\item \textit{Inference.} Given a source input, how to generate a translation sentence from the NMT model?
\item \textit{Learning.} How to effectively learn the parameters of NMT from data?
\end{itemize}

In \ref{sec:overview}, we first describe the basic methods of NMT for addressing the above three questions. We then dive into the details of NMT architectures in \ref{sec:arch} and introduce bidirectional inference and non-autoregressive NMTs in \ref{sec:decoding}. We discuss alternative training objectives and using monolingual data in \ref{sec:objective} and \ref{sec:mono}, respectively.

Despite the great success, NMT is far from perfect. There are several theoretical and practical challenges faced by NMT. We survey the research progress of some important directions. We describe methods for open vocabulary in \ref{sec:vocab}, prior knowledge integration in \ref{sec:knowledge}, and interpretability and robustness in \ref{sec:interpret}.

\begin{figure*}[t]
  \centering
    \includegraphics[width=0.8\textwidth]{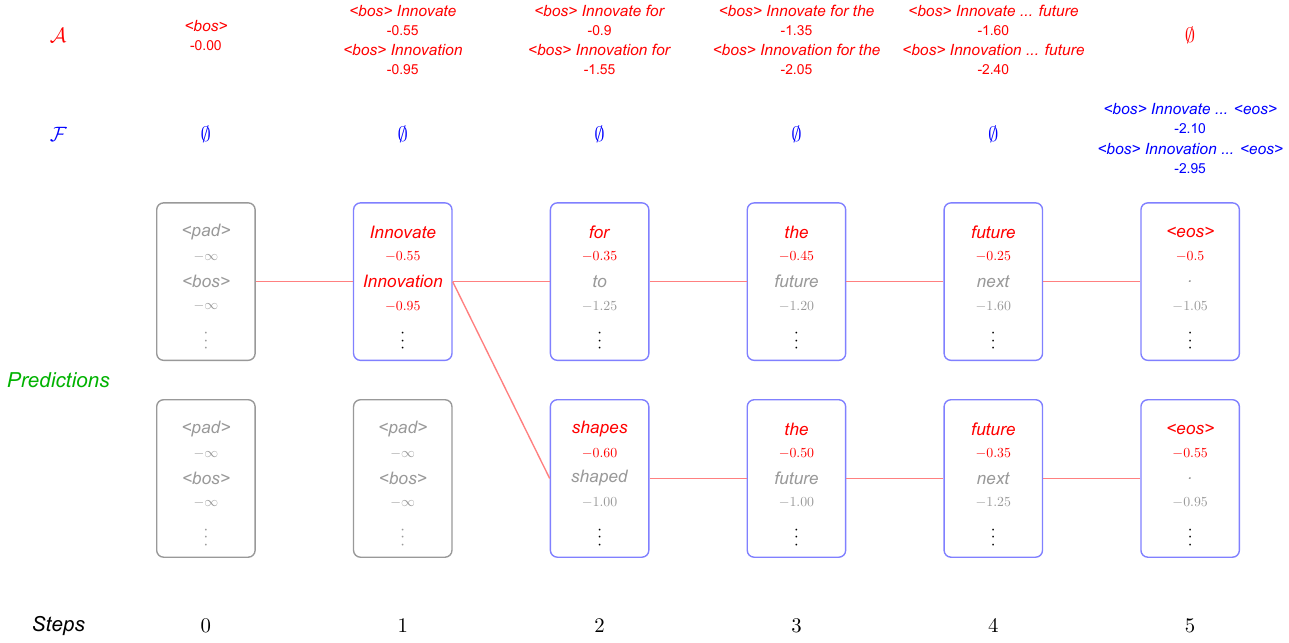}
    \caption{A running example of the beam-search algorithm.}
  \label{fig:beamsearch}
\end{figure*}

\subsection{Overview of NMT} \label{sec:overview}
\subsubsection{Modeling}
Translation can be modeled at different levels, such as document-, paragraph-, and sentence-level. In this article, we focus on sentence-level translation. Besides, we also assume the input and output sentences are sequences. Thus the NMT model can be viewed as a \textit{sequence-to-sequence} model. Assuming we are given a source sentence $\mathbf{x}=\{x_1,\ldots,x_S\}$ and a target sentence $\mathbf{y}=\{y_1,\dots,y_T\}$. By using the chain rule, the conditional distribution can be factorized from left-to-right (L2R) as
\begin{equation}
P(\bm{y}=\mathbf{y}|\bm{x}=\mathbf{x})=\prod_{t=1}^T P(y_t|y_0,\dots,y_{t-1}, x_1, \ldots, x_S).
\label{eq:condprob}
\end{equation}
NMT models which conform the Eq.~(\ref{eq:condprob}) is referred to as \emph{L2R autoregressive} NMT~\cite{kal2013rctm,cho2014encdec,sutskever2014seq2seq,bahdanau2015nmt}, for the prediction at time-step $t$ is taken as a input at time-step $t+1$. 

Almost all neural machine translation models employ the \emph{encoder-decoder framework}~\cite{cho2014encdec}. The encoder-decoder framework consists of four basic components: the embedding layers, the encoder and decoder networks, and the classification layer. Figure~\ref{fig:nmt} shows a typical autoregressive NMT model using the encoder-decoder framework, which we shall use as an example. ``$\texttt{<bos>}$'' and ``\texttt{<eos>}'' are special symbols that mark the beginning and ending of a sentence, respectively.

The embedding layer embodies the concept of \emph{continuous representation}. It maps a discrete symbols $x_t$ into a continuous vector $\mathbf{x}_t \in \mathbb{R}^d$, where $d$ denotes the dimension of the vector. The embeddings are then fed into later layers for more finer-grained feature extraction.

The encoder network maps the source embeddings into hidden continuous representations. To learn expressive representations, the encoder must be able to model the ordering and complex dependencies that existed in the source language. Recurrent neural networks (RNN) are suitable choice for modeling variable-length sequences. With RNNs, the computation involves in encoder can be described as
\begin{equation}
\mathbf{h}_t = \mathrm{RNN}_{\textrm{ENC}}(\mathbf{x}_t, \mathbf{h}_{t-1}).
\end{equation}
By iteratively applying the state transition function $\mathrm{RNN}_{\textrm{ENC}}$ over the input sequence, we can use the final state $\mathbf{h}_S$ as the representation for the entire source sentence, and then feed it to the decoder.

The decoder can be viewed as a language model conditioned on $\mathbf{h}_S$. The decoder network extracts necessary information from the encoder output, and also models the long-distance dependencies between target words. Given the start symbol $y_0=\texttt{<bos>}$ and the initial state $\mathbf{s}_0 = \mathbf{h}_S$, the RNN decoder compresses the decoding history $\{y_0,\ldots,y_{t-1}\}$ into a state vector $\mathbf{s}_t \in \mathbb{R}^{d}$:
\begin{align}
  \mathbf{s}_t = \mathrm{RNN}_{\textrm{DEC}}(\mathbf{y}_{t-1},\mathbf{s}_{t-1}).
 \end{align}

The classification layer predicts the distribution of target tokens. The classification layer is typically a linear layer with \emph{softmax} activation function. Assuming the vocabulary of target language is $\mathcal{V}$, and $|V|$ is the size of the vocabulary. Given an decoder output $\mathbf{s}_t \in \mathbb{R}^d$, the classificaition layer first maps $\mathbf{h}$ to a vector $\mathbf{z}$ in the vocabulary space $\mathbb{R}^{|V|}$ with the linear map. Then the softmax function is used to ensure the output vector is a valid probability:
\begin{align}
   \mathrm{softmax}(\mathbf{z}) = \frac{\exp(\mathbf{z})}{\sum_{i=1}^{|V|}\exp(\mathbf{z}_{[i]})},
\end{align}
where we use $\mathbf{z}_{[i]}$ to denote the $i$-th component in $\mathbf{z}$.


\subsubsection{Inference}

Given an NMT model and a source sentence $\mathbf{x}$, how to generate a translation from the model is an important problem. Ideally, we would like to find the target sentence $\mathbf{y}$ which maximizes the model prediction $P(\mathbf{y}|\bm{x}=\mathbf{x};\bm{\theta})$ as the translation. However, due to the intractably large search space, it is impractical to find the translation with the highest probability. Therefore, NMT typically uses local search algorithms such as \emph{greedy search} or \emph{beam search} to find a local best translation. 

Beam search is a classic local search algorithm which have been widely used in NMT. Previously, beam search have been successfully applied in SMT.
The beam search algorithm keeps track of $k$ states during the inference stage. Each state is a tuple $\langle y_0\ldots y_{t}, v\rangle$, where $y_0\ldots y_{t}$ is a candidate translation, and $v$ is the log-probability of the candidate. At each step, all the successors of all $k$ states are generated, but only the top-$k$ successors are selected. The algorithm usually terminates when the step exceed a pre-defined value or $k$ full translation are found. It should be noted that the beam search will degrade into the greedy search if $k=1$.

\begin{algorithm}[h!]
$t \leftarrow 1$ \;
$\mathcal{A} = \{\langle \mathtt{<}\mathtt{bos}\mathtt{>}, 0 \rangle \}$ ; \algorithmiccomment{\textit{The set of alive candidates}} \\
$\mathcal{F} = \{\}$ ; \algorithmiccomment{\textit{The set of finished candidates}} \\

\While(){$t < \mathtt{max\_length}$} {
  $\mathcal{C} = \{\}$ \;
  \For(){$\langle y_0\ldots y_{t-1}, v \rangle \in \mathcal{A}$}{
    $\bm{p} \leftarrow \mathrm{NMT}(y_0\ldots y_{t-1}, \mathbf{x})$ ; \\

    \For(){$w \in \mathcal{V}$}{
      $y_t \leftarrow w$ \;
      $l \leftarrow \log(\bm{p}[w]) $ \;
      $\mathcal{C} \leftarrow \mathcal{C} \cup \{\langle y_0\ldots y_t, v + l\rangle \}$ \;
    }
  }

  $\mathcal{C} \leftarrow \mathrm{TopK}(\mathcal{C}, k)$ ; \\

  \For{$\langle y_0\ldots y_{t}, v \rangle \in \mathcal{C}$}{
    \uIf{$y_t == \mathtt{<}\mathtt{eos}\mathtt{>}$}{
      $\mathcal{F} \leftarrow \mathcal{F} \cup \{\langle y_0\ldots y_t, v\rangle \}$ \;
    }
    \Else{
      $\mathcal{A} \leftarrow \mathcal{A} \cup \{\langle y_0\ldots y_t, v\rangle \}$ \;
    }
  }

  $\mathcal{A} \leftarrow \mathrm{TopK}(\mathcal{A}, k)$ ; \\
  $\mathcal{F} \leftarrow \mathrm{TopK}(\mathcal{F}, k)$ ; \\
  $t \leftarrow t+1$ \;
}
$\langle y_0\ldots y_t,v \rangle \leftarrow \mathrm{Top}(\mathcal{F})$ \;
\Return $y_1\ldots y_t$
\caption{The beam search algorithm}
\label{alg:beam_search}
\end{algorithm}

The pseudo-codes of the beam search algorithm are given in $\ref{alg:beam_search}$. We also give a running example of the algorithm in Figure~\ref{fig:beamsearch}.

\subsubsection{Training of NMT Models}
NMT typically uses maximum log-likelihood (MLE) as the training objective function, which is a commonly used method of estimating the parameters of a probability distribution. Formally, given the training set $\mathcal{D} = \{ \langle \mathbf{x}^{(s)}, \mathbf{y}^{(s)} \rangle \}_{s=1}^{S}$, the goal of training is to find a set of model parameters that maximize the log-likelihood on the training set:
\begin{align}
\hat{\bm{\theta}}_{\mathrm{MLE}} = \argmax_{\bm{\theta}} \Big\{ \mathcal{L}(\bm{\theta}) \Big\}, \label{MLEloss}
\end{align}
where the log-likelihood is defined as
\begin{align}
\mathcal{L}(\bm{\theta}) = \sum_{s=1}^{S} \log P(\mathbf{y}^{(s)}|\mathbf{x}^{(s)}; \bm{\theta}).
\end{align}

By the virtue of \emph{back-propagation} algorithm, we can efficiently compute the gradient of $\mathcal{L}$ with respect to $\bm{\theta}$. The training of NMT models usually adopts \emph{stochastic gradient search} (SGD) algorithm. Instead of computing gradients on the full training set, SGD computes the loss function and graidents on a \emph{minibatch} of the training set. The plain SGD optimzier updates the parameters of an NMT model with the following rule:

\begin{align}
\bm{\theta} \leftarrow \bm{\theta} - \alpha \nabla_{\bm{\theta}} \mathcal{L}(\bm{\theta}),
\end{align}
where $\alpha$ is the \emph{learning rate}. With well-chosen learning rate, the parameters of NMT are guaranteed to converge into a local optima. In practice, instead of plain SGD optimizer, adaptive learning rate optimizers such as Adam~\cite{kingma2014adam} are found to greatly reduce the training time.

\subsection{Architectures} \label{sec:arch}

\subsubsection{Evolution of NMT Architectures} \label{sec:fund-arch}
Since 2013, there are attempts to build a pure neural MT. Early NMT architectures such as RCTM~\cite{kal2013rctm}, RNNEncdec~\cite{cho2014encdec}, and Seq2Seq~\cite{sutskever2014seq2seq} adopt a \emph{fixed-length} approach, where the size of source representation is fixed regardless the length of source sentences. These works typically use recurrent neural networks (RNN) as the decoder network for generating variable-length translation. However, it is found that the performance of this approach degrades as the length of the input sentence increases~\cite{cho2014nmt}. Two explanations can account for this phenomenon:
\begin{enumerate}
\item The fixed-length representations have become the bottleneck during the encoding process for long sentences~\cite{cho2014encdec}. As the encoder is forced to compress the entire source sentence into a set of fixed-length vectors, some important information may be lost in this process.
\item The longest path between the source words and target words is $O(S+T)$, and it is challenging for neural networks to learn long-term dependencies~\cite{bengio1994long}. \citet{sutskever2014seq2seq} found that reverse the source sentence can significantly improve the performance of the fixed-length approach. By reversing the source sentence, the paths between the beginning words of source and target sentences are reduced, thus the optimization problem becomes easier.
\end{enumerate}

\begin{figure}[t]
  \centering
  \includegraphics[width=0.4\textwidth]{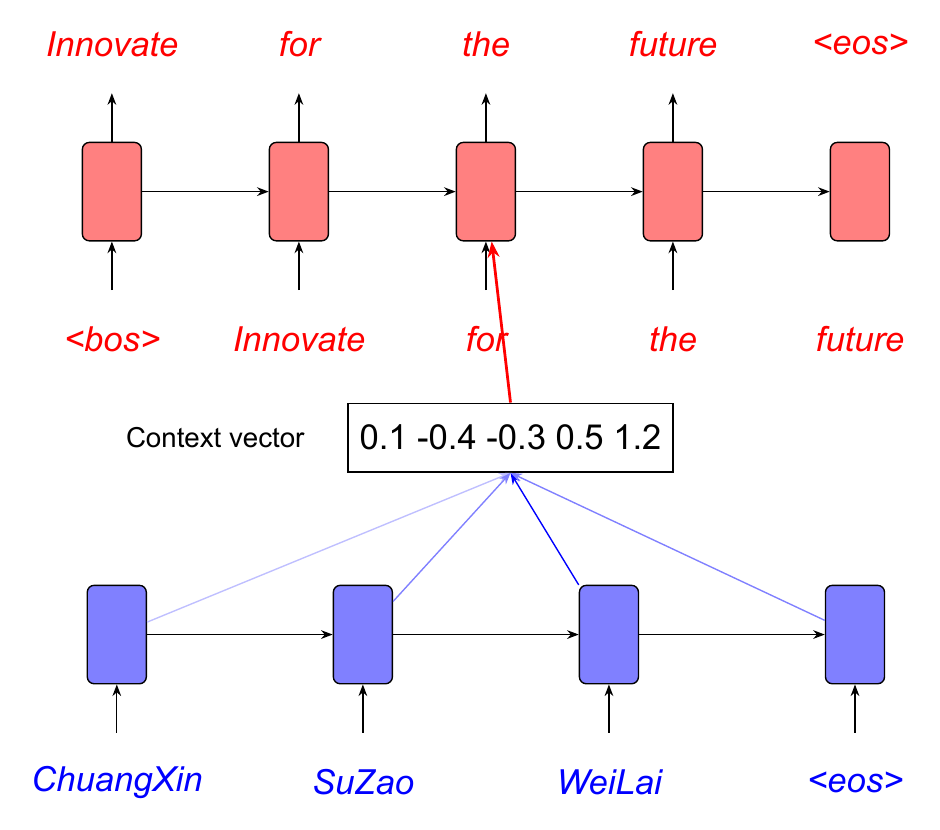}
  \caption{At each decoding step, the attention mechanism dynamically generates a context vector based on the most relevant source representations for predicting the next target word.}\label{fig:encdec-comp}
\end{figure}

Due to these limitations, later NMT architectures switch to \textit{variable-length} source representations, where the length of source representations depends on the length of the source sentence. The RNNsearch architecture~\cite{bahdanau2015nmt} introduces \emph{attention mechanism}, which is an important approach to implementing variable-length representations. Figure~\ref{fig:encdec-comp} shows the comparison between fixed-length and variable-length approaches. By using the attention mechanism, the paths between any source and target words are within a constant length. As a result, the attention mechanism has eased optimization difficulty.

With the breakthrough of deep learning, NMT with deep neural networks have attracted much research interest. Seq2Seq~\cite{sutskever2014seq2seq} is the first architecture demonstrate the potential of deep NMT. Later architectures such as GNMT~\cite{wu2016gnmt}, ByteNet~\cite{kal2016bytenet}, ConvSeq-2Seq~\cite{gehring2017conv}, and Transformer~\cite{vaswani2017attention} all use multi-layered neural networks. ByteNet and ConvSeq2Seq have replaced RNNs with convolutional neural networks (CNN) in their architectures while Transformer relies entirely on self-attention networks (SAN). Both CNNs and SANs can reduce the sequential operations invovled in RNNs, and benefit from the parallel computation provided by modern devices such as GPU or TPU. Importantly, SAN can further reduce the longest path between two target tokens. We shall later describe the techniques for building these architectures.

\subsubsection{Attention Mechanism} \label{sec:attn}
\begin{figure}[t!]
  \centering
  \begin{subfigure}[b]{0.3\textwidth}
    \centering
    \includegraphics[width=\textwidth]{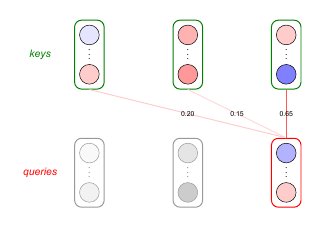}
    \caption{Given a query vector and key vectors, the attention network first computes a weight vector through the scoring function.}\label{fig:encdec}
  \end{subfigure}
  \\
  \begin{subfigure}[b]{0.3\textwidth}
    \centering
    \includegraphics[width=\textwidth]{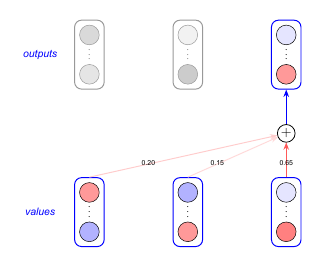}
    \caption{Each output vector is computed as a weighted sum of value vectors.}\label{fig:attn}
  \end{subfigure}
  \caption{Detailed computations involved in the attention mechanism.}\label{fig:att}
\end{figure}

The introduction of attention mechanism~\cite{bahdanau2015nmt} is a milestone in NMT architecture research. The attention network computes the relevance of each value vector based on queries and keys. This can also be interpreted as a content-based addressing sche-me~\cite{graves2014neural}. Formally, given a set of $m$ query vectors $\mathbf{Q} \in \mathbb{R}^{m\times d}$, a set of $n$ key vectors $\mathbf{K} \in \mathbb{R}^{n \times d}$ and associated value vectors $\mathbf{V} \in \mathbb{R}^{n \times d}$, the computation of attention network involves two steps. The first step is to compute the relevance between keys and values, which is formally described as
\begin{equation}
\mathbf{R} = \mathrm{score}(\mathbf{Q},\mathbf{K}),
\end{equation}
where $\mathrm{score}(\cdot)$ is a scoring function which have several alternatives. $\mathbf{R} \in \mathbb{R}^{m\times n}$ is a matrix storing the relevance score between each keys and values. The next step is compute the output vectors. For each query vector, the corresponding output vector is expressed as a weighted sum of value vectors:
\begin{align}
\mathrm{Attention}(\mathbf{Q},\mathbf{K},\mathbf{V})&=\mathrm{softmax}(\mathbf{R})\cdot\mathbf{V}.
\end{align}
Figure~\ref{fig:att} depicts the two steps involved in the computation of attention mechanism.

Considering on the scoring function, the attention networks can be roughly classified into two categories: additive attention~\cite{bahdanau2015nmt} and dot-product attention~\cite{luong2015effective}. The additive attention models $\mathrm{score}$ through a feed-forward neural network:
\begin{equation}
\mathbf{R}_{[i,j]} = \mathbf{v}^{\top} \tanh(\mathbf{W}_s \mathbf{Q}_{[i]}+\mathbf{U}_s\mathbf{K}_{[j]}),
\end{equation}
where $\mathbf{W}_s \in \mathbb{R}^{d \times d}, \mathbf{U}_s \in \mathbb{R}^{d \times d}$, and $\mathbf{v} \in \mathbb{R}^{d\times 1}$ are learnable parameters. On the other hand, the dot-product attention uses dot production to compute the matching score:
\begin{equation}
\mathbf{R}_{[i,j]} = \mathbf{Q}_{[i]}^{\top} \mathbf{K}_{[j]}.
\end{equation}

In practice, the dot-product attention is much faster than the additive attention. However, the dot-product attention is found to be less stable than the additive attention when $d$ is large~\cite{vaswani2017attention}. \citet{vaswani2017attention} suspect that the dot-products grow large in magnitude for large values of $d$, which may resulting extremely small gradients caused by the softmax function. To remedy this issue, they propose to scale the dot-products by $\frac{1}{\sqrt{d}}$.

The attention mechanism is usually used as a part of the decoder network. Another type of attention network called self-attention network, is widely used in both the encoder and decoder of NMT. We shall describe self-attention and other variants of attention network later.

\subsubsection{RNNs, CNNs, and SANs}\label{sec:enc-dec}
There are many methods of building powerful encoders and decoders, which can roughly divide into three categories: the recurrent neural network (RNN) based methods, convolutional neural network (CNN) based methods, and self-attention network (SAN) based methods. There are several aspects we need to take into considerations for building an encoder and decoder:
\begin{enumerate}
\item \emph{Receptive field}. We hope each output produced by the encoder and decoder can potentially encode arbitrary information in the input sequence.
\item \emph{Computational complexity}. It is desirable to a use network with lower computational complexity.
\item \emph{Sequential operations}. Too many sequential operations preclude the parallel computation within the sequence.
\item \emph{Position awareness}. The network should distinguish the ordering presents in the sequence.
\end{enumerate}
Table~\ref{tab:nn} summarizes the computation as well as the above-mentioned aspects of typical RNN, CNN, and SAN.

\begin{table*}[t]
\centering
\resizebox{0.6\textwidth}{!}{
\begin{tabular}{l|c|c|c|c|c}
\toprule
{\bf Layer} & {\bf Computation} & {\bf R.F.} & {\bf Complexity} & {\bf S.O.} & {\bf P.A.} \\ \hline
RNN & $\mathbf{h}_{l,t} = \mathbf{W}\mathbf{h}_{l-1,t} + \mathbf{U}\mathbf{h}_{l,t-1} $ & $\infty$ & $O(n\cdot d^2)$ & $O(n)$ & Yes \\
CNN & $\mathbf{h}_{l,t} = \sum_{i=1}^k \mathbf{W}^{(i)}\mathbf{h}_{l-1,t+i-\lceil \frac{k+1}{2} \rceil} $ & $k$ & $O(k\cdot n \cdot d^2)$ & $O(1)$ & Yes \\
SAN & $\mathbf{h}_{l,t} = \sum_{i=1}^n\alpha_{l,i}\mathbf{h}_{l-1,i} $ & $\infty$ & $O(n^2\cdot d)$ & $O(1)$ & No  \\
\bottomrule
\end{tabular}}
\caption{Comparisons between different neural network layers. We use R.F. to denote the receptive field, S.O. to denote the number of sequential operations, and P.A. to denote the position awareness of the layer. $t$ is the position in the sequence, $l$ is the layer number. For CNN, $k$ is the filter width and $\mathbf{W}^{(i)}$ is the weight of the $i$-th filter.  }\label{tab:nn}
\end{table*}

\begin{figure}[!th]
  \centering
     \begin{subfigure}[b]{0.2\textwidth}
         \centering
         \includegraphics[width=\textwidth]{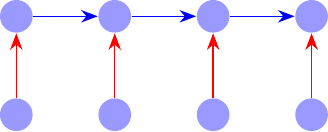}
         \caption{RNN.}
     \end{subfigure}
     \\
     \begin{subfigure}[b]{0.2\textwidth}
         \centering
         \includegraphics[width=\textwidth]{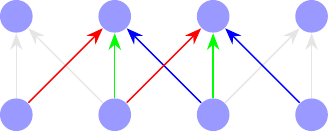}
         \caption{CNN with kernel width $k=3$.}
     \end{subfigure}
     \\
     \begin{subfigure}[b]{0.2\textwidth}
         \centering
         \includegraphics[width=\textwidth]{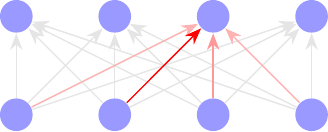}
         \caption{SAN.}
     \end{subfigure}
        \caption{Overview of the computation diagram of RNN, CNN, and SAN. To be clarity, we use a node to denote the input or output vector of a specific layer.}
  \label{fig:nn}
\end{figure}

\begin{figure}[!ht]
  \centering
     \begin{subfigure}[b]{0.259\textwidth}
         \centering
         \includegraphics[width=\textwidth]{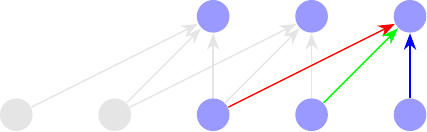}
         \caption{CNN with additional zero paddings.}
     \end{subfigure}
     \\
     \begin{subfigure}[b]{0.2\textwidth}
         \centering
         \includegraphics[width=\textwidth]{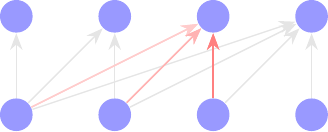}
         \caption{SAN with causal masking.}
     \end{subfigure}
  \caption{The computation of CNN and SAN during decoding.}
  \label{fig:decnn}
\end{figure}

Figure~\ref{fig:nn} gives an overview of the ways of RNN, CNN, and SAN to encode sequences, respectively. In order to keep the auto-regressive property of NMT decoder during training, CNN and SAN furthur needs additional padding and masking to prevent the network from seeing future words. Figure~\ref{fig:decnn} shows padding and masking used in CNN and SAN.

As we can see in Figure~\ref{fig:nn}(a), RNNs are a family of sequential models that repeatedly apply the same state transition function to sequences. In theory, RNNs are among the most powerful family of neural networks~\cite{siegelmann1995computational}. However, it suffers from severe vanishing and exploding gradient problem~\cite{bengio1994long} in practice. RNNs with gates, such as long short-term memory (LSTM)~\cite{hochreiter1997lstm} and gated recurrent unit (GRU)~\cite{cho2014encdec} have been proposed to alleviate this problem. Another way to stabilize the training is to incorporate normalization layers, such as layer normalization~\cite{ba2016layer}.


\begin{figure}[!ht]
  \centering
     \begin{subfigure}[b]{0.2\textwidth}
         \centering
         \includegraphics[width=\textwidth]{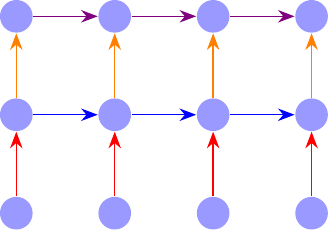}
         \caption{Deep RNN.}
     \end{subfigure}
     \\
     \begin{subfigure}[b]{0.227\textwidth}
         \centering
         \includegraphics[width=\textwidth]{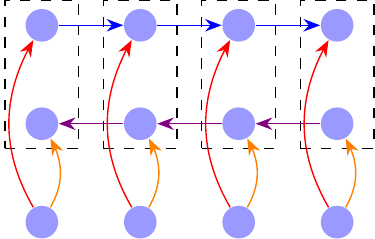}
         \caption{Bidirectional RNN.}
     \end{subfigure}
     \\
     \begin{subfigure}[b]{0.2\textwidth}
         \centering
         \includegraphics[width=\textwidth]{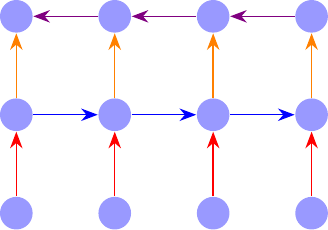}
         \caption{Alternating RNN.}
     \end{subfigure}
        \caption{Three extensions to RNNs.}
  \label{fig:rnn}
\end{figure}

Figure~\ref{fig:rnn} shows three extensions to RNNs that are widely used in NMT literature. Deep RNNs is one important way to increase the expressive power of RNNs. However, training deep neural networks is challenging because it also faces the vanishing and exploding gradient problem. There are many ways to construct deep RNNs, and the most popular one is by stacking multiple RNNs with \emph{residual connections}~\cite{he2016deep}. The residual connection is an important method to construct deep neural networks. Residual connections use the identity mappings as the skip connections, which is formally described as
\begin{equation}
\mathbf{y} = \mathbf{x} + f(\mathbf{x}),
\end{equation}
where $\mathbf{x}$, and $\mathbf{y}$ are input and output, respectively. $f$ is the neural network. By using identity mappings, the gradient signal can directly propagate into lower layers. Bidirectional RNNs~\cite{bahdanau2015nmt} use two RNNs to process the same sequence in opposite directions, and concatenating the results of both RNNs to be the final output. In this way, each output of bidirectional RNNs encodes all the tokens in the sequence. An alternative to bidirectional RNNs is alternating RNNs~\cite{zhou2015end}, which consists of RNNs in opposite directions in adjacent layers.

Besides the difficulty training of RNNs, another major drawback of RNNs is that RNNs are sequential models in nature, which cannot benefit from the parallel computations provided by modern GPUs. CNNs and SANs, however, which fully exploit the parallel computation within sequences, are widely used in newer NMT architectures.

Convolutional neural network (CNN) was first introduced into NMT in 2013~\cite{kal2013rctm}. However, it was not as successful as RNNs until 2017~\cite{gehring2017conv}. The main obstacle for applying CNNs is its limited receptive field. Stacking $L$ CNNs with kernel width $k$ can increase the receptive field from $k$ to $L\cdot(k-1) + 1$. The network needs to go deeper with large $L$ and adopt large kernel size $k$ to model long sentences. However, learning deep CNNs is challenging, and using large kernel size $k$ may significantly increase the complexity and parameters involved in CNNs.

\begin{figure}[th]
  \centering
     \begin{subfigure}[b]{0.259\textwidth}
         \centering
         \includegraphics[width=\textwidth]{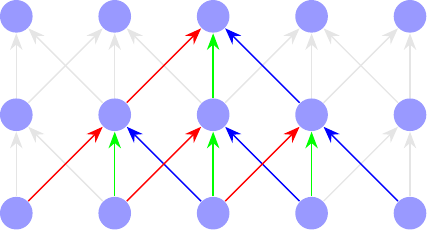}
         \caption{}
     \end{subfigure}
     \hfill
     \begin{subfigure}[b]{0.380\textwidth}
         \centering
         \includegraphics[width=\textwidth]{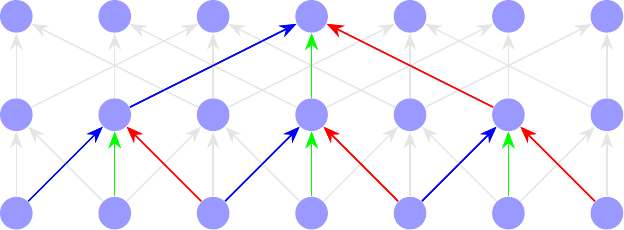}
         \caption{}
     \end{subfigure}
  \caption{Comparison between CNN and dilated CNN. (a) Two layers CNN with filter width $k=3$ for each layer. (b) Dilated CNN with filter width $k=3$ for all layers, dilation rate $r=1$ in layer 1 and $r=2$ in layer 2.}
  \label{fig:cnn}
\end{figure}

One solution to increase the receptive field without using a large $k$ is through dilation~\cite{kal2016bytenet}. Figure~\ref{fig:cnn} shows the comparison between plain CNN and dilated CNN. Plain CNN can be viewed as a special case of dilated CNN with a dilation rate $r=1$. The computation of dilated CNN is mathematically formulated as
\begin{equation}
\mathbf{h}_{l,t} = \sum_{i=1}^k \mathbf{W}^{(i)}\mathbf{h}_{l-1,t+(i-\lceil \frac{k+1}{2} \rceil)\times r} .
\end{equation}
Stacking $L$ dilated CNNs whereby the dilation rates are doubled every layer, the receptive field increases to $(2^L-1)\cdot(k-1) + 1$. As a result, the receptive field grows exponentially with $L$, as opposed to linearly with $L$ in plain CNN.

\begin{figure}[t]
 \centering
 \begin{subfigure}[b]{0.3\textwidth}
         \centering
         \includegraphics[width=\textwidth]{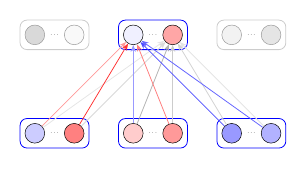}
         \caption{}
     \end{subfigure}
     \\
     \begin{subfigure}[b]{0.3\textwidth}
         \centering
         \includegraphics[width=\textwidth]{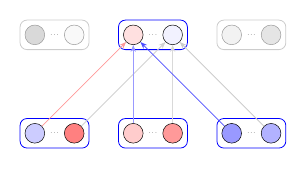}
         \caption{}
     \end{subfigure}
 \caption{Comparison between CNN and depthwise CNN. Each node in the graph represents a neuron instead of a vector. (a) Plain CNN. We highlight the computation of the first neuron in the output vector. (b) Depthwise CNN. Note that the connections are significantly reduced compared with plain CNN.}
 \label{fig:lcnn}
\end{figure}

Another solution is to reduce the computations involved in CNN. Depthwise convolution~\cite{kaiser2017slicenet} reduces the complexity from $O(kd^2)$ to $O(kd)$ by performing convolution independently over channels. Figure~\ref{fig:lcnn} depicts the comparison between CNN and depthwise CNN. The output of the depthwise convolution layer is defined as
\begin{equation}
\mathbf{h}_{l,t} = \sum_{i=1}^k \mathbf{w}^{(i)} \odot \mathbf{h}_{l-1,t+i-\lceil \frac{k+1}{2} \rceil},
\end{equation}
where $\mathbf{w}^{(i)}$ is the $i$-th column of weight matrix $\mathbf{W} \in \mathbb{R}^{k\times d}$. Lightweight convolution~\cite{wu2019pay} further reduces the number of parameters of depthwise convolution through weight sharing.

\begin{figure}[th]
  \centering
  \includegraphics[width=0.3\textwidth]{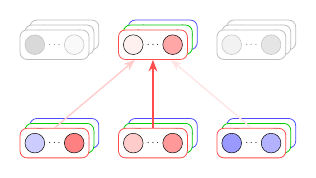}
  \caption{An illustration of multi-head attention.}\label{fig:mha}
\end{figure}

Self-attention network (SAN)~\cite{vaswani2017attention} is a special case of attention network where the queries, keys, and values come from the same sequence. Similar to CNN, SAN is trivial to parallelize. Furthermore, Each output in SAN also has infinite receptive fields, which is the same with RNN. In SAN, the queries, keys, and values are typically obtained through a linear map of the input representations. The scaled dot-product self-attention mechanism can be formally described as
\begin{equation}
\mathrm{Attention}(\mathbf{Q},\mathbf{K},\mathbf{V})=\mathrm{softmax}(\frac{\mathbf{Q}\mathbf{K}^{\top}}{\sqrt{d}})\mathbf{V}.
\end{equation}

Multi-head attention~\cite{vaswani2017attention} is an extended attention network with multiple parallel heads. Each head attends information from different subspace across value vectors. As a result, multi-head attention can perform more flexible transformations than the single-head attention. We give an illustration of multi-head attention in Figure~\ref{fig:mha}.

The major disadvantage of SAN network is that it ignores the ordering of words in the sequence. To remedy this, SAN needs additional position encoding to differentiate orders. \citet{vaswani2017attention} proposed a sinusoid style position encoding, which is formulated as
\begin{align}
\textrm{timing}(t, 2i) &= \sin(t / 10000^{2i / d}), \\
\textrm{timing}(t, 2i + 1) &= \cos(t / 10000^{2i / d}),
\end{align}
where $t$ is the position and $i$ is the dimension index. Another popular way of position encoding is to learn an additional position embedding. Finally, the position encoding is added to each word representation, so the same words with different positions can have different representations.

\begin{figure}[th]
  \centering
  \includegraphics[width=0.35\textwidth]{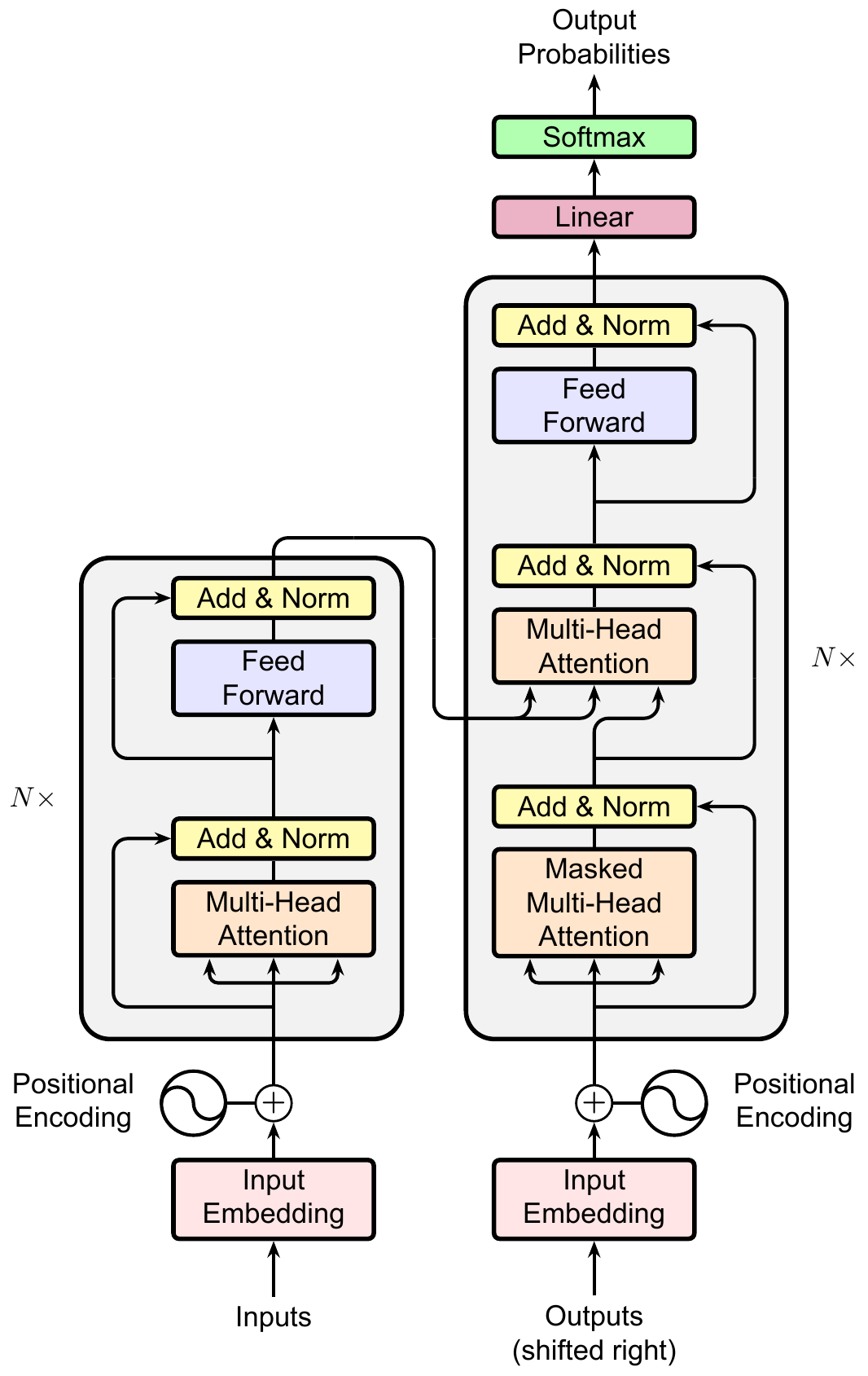}
  \caption{The Transformer architecture.}\label{fig:transformer}
\end{figure}

\subsubsection{Comparison of Fundamental Architectures}

We take the state-of-the-art Transformer architecture~\cite{vaswani2017attention} as an example to put all things together. Figure~\ref{fig:transformer} shows the architecture of Transformer. The Transformer model relies solely on attention networks, with additional sinusoid-style position encoding added to input embedding. The Transformer network consists of a stack of 6 encoder layers and 6 decoder layers. Each encoder layer contains two sub-layers whereas each decoder layer contains three sub-layers. To stabilize optimization, Transformer uses residual connection and layer normalization in each sub-layer.

\begin{table*}[!t]
  \centering
  \resizebox{0.8\textwidth}{!}{
  \begin{tabular}{c | c | c | c | c | c | c}
  \toprule
    {\bf Model} & {\bf Encoder} & {\bf Decoder} & {\bf Complexity} & {\bf V.R.} & {\bf Path$_{\textrm{E}}$} & {\bf Path$_\textrm{D}$}\\ \hline
    \textsc{RCTM 1}~\cite{kal2013rctm} & CNN & RNN & $O(S^2 + T)$ & No & $S$ & $T$ \\
    \textsc{RCTM 2}~\cite{kal2013rctm} & CNN & RNN & $O(S^2 + T)$ & Yes & $S$ & $T$ \\
    \textsc{RNNEncdec/Seq2Seq}~\cite{cho2014encdec,sutskever2014seq2seq} & RNN & RNN & $O(S + T)$ & No & $S + T$ & $T$ \\
    \textsc{RNNsearch}~\cite{bahdanau2015nmt} & RNN & RNN & $O(ST)$ & Yes & $1$ & $T$ \\
    \textsc{ByteNet}~\cite{kal2016bytenet} & CNN & CNN & $O(S+T)$ & Yes & $c$ & $c$ \\
    \textsc{ConvSeq2Seq}~\cite{gehring2017conv} & CNN & CNN & $O(ST)$ & Yes & $1$ & $c$ \\
    \textsc{Transformer}~\cite{vaswani2017attention} & SAN & SAN & $O(S^2+ST+T^2)$ & Yes & $1$ & $1$ \\\bottomrule
  \end{tabular}}
  \caption{Comparison of fundamental architectures. V.R. denotes whether the architecture employs variable representation. Path$_{\textrm{E}}$ denotes the longest path between the source and target tokens. Path$_{\textrm{D}}$ denotes the longest path between two target tokens.}\label{tab:arch}
\end{table*}

We summarize the comparison of fundamental NMT architectures in Table~\ref{tab:arch}. We highlight several important aspects of these fundamental architectures.

\begin{figure}[t]
 \centering
 \begin{subfigure}[b]{0.48\textwidth}
         \centering
         \includegraphics[width=\textwidth]{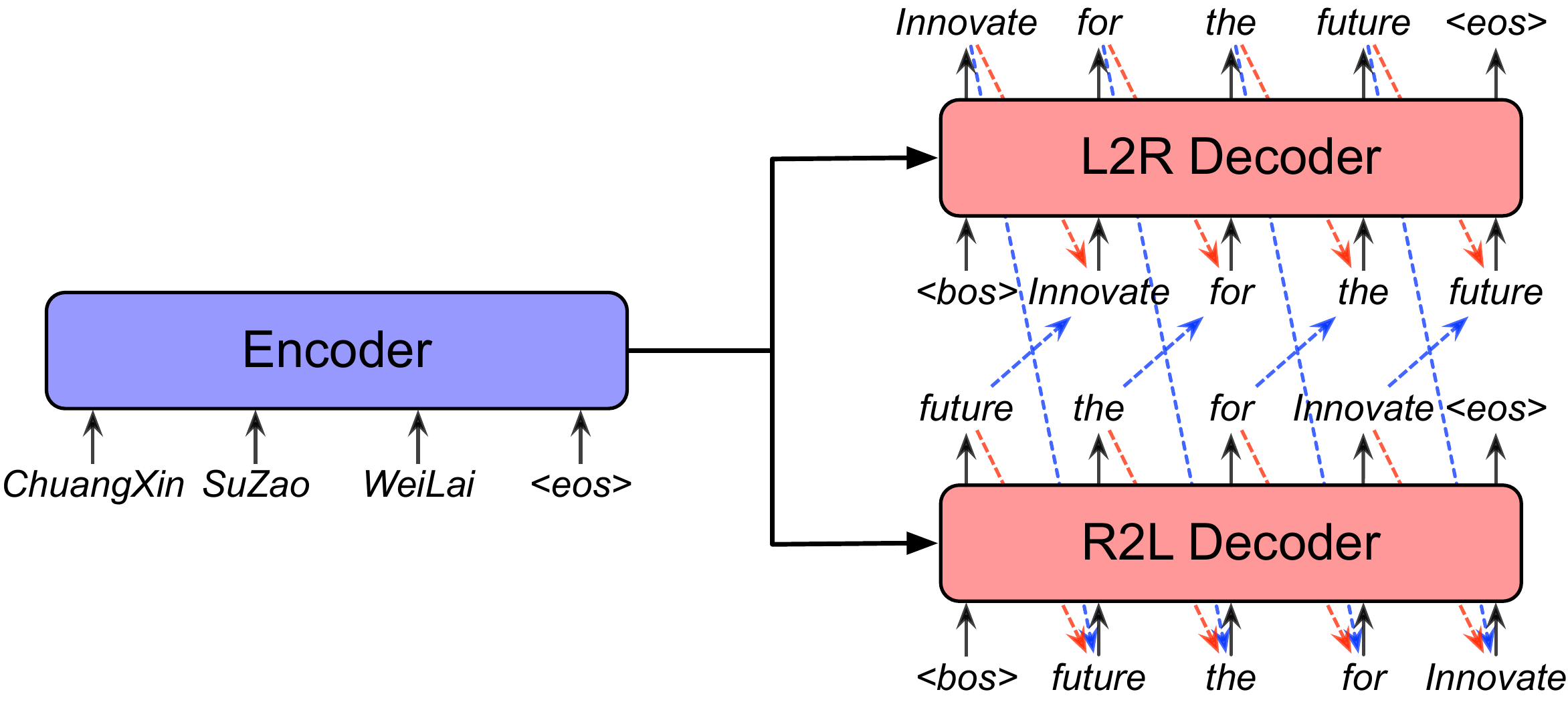}
         \caption{Bidirectional inference.}
         \label{fig:bi_dec}
     \end{subfigure}
     \\
 \begin{subfigure}[b]{0.48\textwidth}
         \centering
         \includegraphics[width=\textwidth]{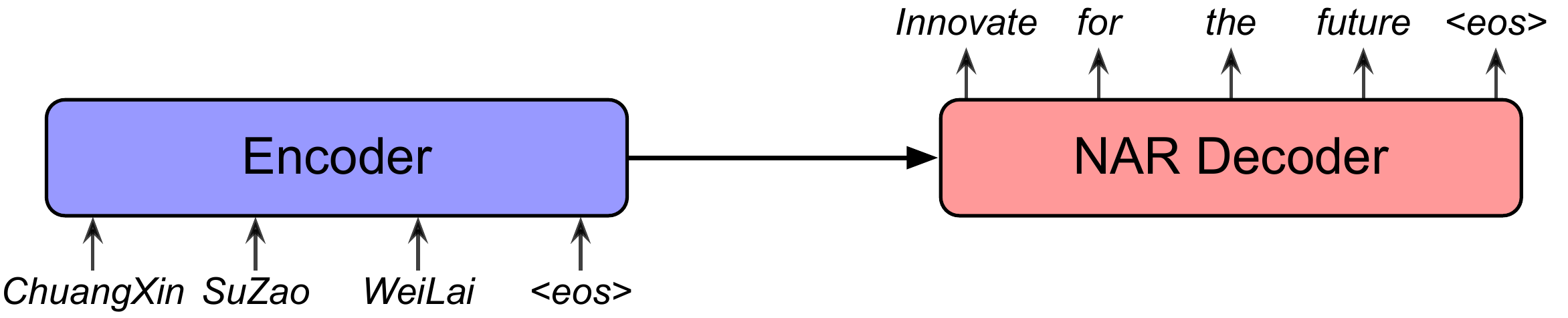}
         \caption{Non-autoregressive decoding.}
         \label{fig:nar_dec}
     \end{subfigure}
 \caption{Comparisons of different decoding strategies. (a) Bidirectional decoding: generates a sentence in both left-to-right (L2R) and right-to-left (R2L) directions; (b) Non-autoregressive (NAR) decoding: generates a sentence at one time.}
 \label{fig:dec_method}
\end{figure}

\subsection{Bidirectional Inference and Non-autoregressive NMT}\label{sec:decoding}

The dominate approach to NMT factorizes the conditional probability $P(\mathbf{y}|\mathbf{x})$ from left to right (L2R) auto-repressively. However, the factorization of the distribution is not unique. Researchers~\cite{liu2016agreement,hoang2017towards,zhang2018asynchronous,zhou2019synchronous} have found that models with right-to-left (R2L) factorization are complementary to L2R models. The bidirectional inference is an approach to simultaneously generating translation with both L2R and R2L decoders. In addition to auto-regressive approaches where each output word on previously generated outputs, non-autoregressive NMTs~\cite{gu2018nat} avoids this auto-regressive property and produces outputs in parallel, allowing much lower latency during inference.

\subsubsection{Bidirectional Inference}
Ignoring the future context is another obvious weakness of AR decoding. Thus, a natural idea is that the quality of translation will be improved if autoregressive models can ``know'' the future information. From this perspective, many approaches have been proposed to improve translation performance by exploring the future context. Some researchers proposed to model both past and future context \cite{zheng2018modeling,zheng2019dynamic,zhang2019future} and some others also found that L2R and R2L autoregressive models can generate complementary translations \cite{liu2016agreement,hoang2017towards,zhang2018asynchronous,zhou2019synchronous}. For instance, \citet{zhou2019synchronous} analyzed the translation accuracy of the first and last 4 tokens for L2R and R2L models, respectively. The statistical results show that, in Chinese-English translation, L2R performs better in the first 4 tokens while R2L translates better in the last 4 tokens.

Based on the findings mentioned above, a number of methods have been proposed to combine the advantages of L2R and R2L decoding. These approaches are collectively referred to as bidirectional decoding. Bidirectional decoding based methods can be mainly fall into four categories \cite{zhang2020neural}: (1) agreement between L2R and R2L \cite{liu2016agreement,yang2018regularizing,zhang2019regularizing}, (2) rescore with bidirectional decoding \cite{liu2016agreement,sennrich2016edinburgh}, (3) asynchronous bidirectional decoding \cite{zhang2018asynchronous,su2019exploiting}, and (4) synchronous bidirectional decoding \cite{zhou2019synchronous,zhou2019sequence,zhang2020synchronous}.

Mathematically, the L2R translation order is rather arbitrary, and other arrangements such as R2L factorization are equally correct:
\begin{equation}
  P(\mathbf{y}|\mathbf{x}) = \underbrace{\prod\limits_{t=1}^{T}P(y_t|\mathbf{y}_{<t},\mathbf{x})}_{\text{L2R model}} = \underbrace{\prod\limits_{t=1}^{T}P(y_t|\mathbf{y}_{>t},\mathbf{x})}_{\text{R2L model}}.
\end{equation}
Based on this theoretical assumption, \citet{liu2016agreement,yang2018regularizing}, and \citet{zhang2019regularizing} proposed joint training schemes in which each direction is used as a regularizer for the other direction. Empirical results show that these methods can lead to significant improvements compared with standard L2R and R2L models.

Another common scheme to combine L2R and R2L translations is rescoring (also known as reranking). A strong L2R model firstly produces an $n$-best list of translations, and then an R2L model rescores each translation in the $n$-best list~\cite{liu2016agreement,sennrich2016edinburgh,sennrich2017university}. As the scores from L2R and R2L directions are based on complementary models, the quality of translation can be improved by rescoring.
Recently, \citet{zhang2018asynchronous} introduced a new strategy to exploit both L2R and R2L models. They named this method asynchronous bidirectional decoding (ASBD), which first produces outputs (hidden states) by an R2L model and then uses these outputs to optimize the L2R model. ASBD can be done in three steps: The first step is to train a R2L model with bilingual corpora. The second step is to obtain outputs for each given source sentence using the trained R2L model. Finally, the output of R2L model is used as the additional context with the training data to train the L2R model. Thanks to incorporating the future information from the R2L model, the performance of L2R model can be substantially improved.

Although ASBD improves the quality of translation, it also incurs other problems. The L2R and R2L models are trained separately so that they have no chance to interact with each other. Besides, the L2R model translates source sentences based on the outputs of an R2L model, this degrades the efficiency of inference. To address these problems, \citet{zhou2019synchronous} further proposed a synchronous bidirectional decoding (SBD) method which generates translations using both L2R and R2L inference synchronously and interactively. Specifically, SBD uses a new synchronous attention model to allow both L2R and R2L models ``communicating'' with each other. As shown in Figure~\ref{fig:bi_dec}, the dotted arrows illustrate interactions between L2R and R2L decoding. \citet{zhou2019synchronous} also designed a variant of the standard beam search algorithm to hold L2R and R2L decoding concurrently. The idea behind this algorithm is to maintain that each half beam contains L2R and R2L predictions, respectively. Empirical results show that SBD can significantly improve performance with a slight cost to decoding speed.

\citet{mehri2018middle} proposed a novel middle-out decoder architecture that begins from an initial middle-word and simultaneously expands the sequence in both L2R and R2L directions. \citet{zhou2019sequence} also proposed a similar method that allows L2R and R2L inferences to start concurrently from the left and right sides, respectively. Both L2R and R2L inferences terminate at the middle position. Extensive experiments demonstrate that this method can improve not only the accuracy of translation but also decoding efficiency.

\subsubsection{Non-autoregressive NMTs}
To reduce the latency during inference, \citet{gu2018nat} first proposed the non-autoregressive NMT (NAT) to generate the target words in parallel. Formally, given the source sentence $\mathbf{x}$, the probability of the target sentence $\mathbf{y}$ is modeled as follows:
\begin{equation}
  P_{\mathcal{NA}}(\mathbf{y}|\mathbf{x};\bm{\theta}) = P_{L}(T|\mathbf{x};\bm{\theta}) \cdot \prod\limits_{t=1}^{T}P(y_{t}|\mathbf{x};\bm{\theta}),
  \label{eq:nat}
\end{equation}
where $P_{\mathcal{NA}}(\mathbf{y}|\mathbf{x};\bm{\theta})$ is the NAT model, $P_{L}(T|\mathbf{x};\bm{\theta})$ is a length sub-model to determine the length of target sentence, and $\bm{\theta}$ denotes the set of model parameters.

How to predict the length of target sentence (i.e., $P_{L}(T|\mathbf{x};\bm{\theta})$ in Eq.~(\ref{eq:nat})) is critical for NAT. \citet{gu2018nat} proposed a fertility predictor to predict the length of translation. The fertility of a word in the source side determines how many target words it is aligned to. The fertility predictor can be denoted as
\begin{equation}
  P_{F}(\mathbf{f}|\mathbf{x};\bm{\theta}) = \prod\limits_{s=1}^{S}P(f_{s}|\mathbf{x};\bm{\theta}),
  \label{eq:fertility}
\end{equation}
where $\mathbf{f} = \{f_1, \cdots, f_S\}$ is the fertility of the source sentence that consists of $S$ words, and $\bm{\theta}$ is the set of parameters. At the training phase, the gold fertility of each sentence pair in the training data can be obtained by a word alignment system. At the inference phase, the length of the target sentence can be determined by the fertility predictor:
\begin{align}
  \hat{T} &= \sum\limits_{s=1}^{S}\hat{f_s}, \\
  \hat{f_s} &= \mathop{\rm argmax}\limits_{f_s}P(f_s|\mathbf{x};\hat{\bm{\theta}}),
\end{align}
where $\hat{T}$ is the number of words in the translation of the source sentence $\mathbf{x}$, and $\hat{\bm{\theta}}$ is the set of learned parameters.

Different from autoregressive NMT models that take the previous words (ie., $\mathbf{y}_{<t}$) as the input to predict the next target word $y_t$, NAT lacks such history information. \citet{gu2018nat} also noticed that missing the input of the decoder can greatly impair translation quality. Thus, the authors proposed to copy each source token to the decoder, and the times each input token to be copied is its ``feritility''. \citet{gu2018nat} also used knowledge distillation~\cite{kim2016sequence}, which employs strong autoregressive models as the ``teachers'' to improve the performance. Knowledge distillation has proven necessary for non-autoregressive translation \cite{zhou2019understanding,gu2018nat,lee2018deterministic,libovicky2018end,ghazvininejad2019mask}.

Despite the promising success of NAT, which can boost the decoding efficiency by about 15 times speedup compared with vanilla Transformer, NAT suffers from considerable quality degradation. Recently, many methods have been proposed to narrow the performance gap between non-autoregressive NMT and autoregressive NMT~\cite{lee2018deterministic,wang2018semi,guo2019non,shao2019retrieving,wang2019non,stern2019insertion,wei2019imitation,akoury2019syntactically,ghazvininejad2019mask,gu2019insertion}.

To take advantage of both autoregressive NMT and non-autoregressive NMT, \citet{wang2018semi} designed a semi-autoregressive Transformer (SAT) model. SAT keeps the autoregressive property in global but performs parallel translation in local. Specifically, SAT produces $K$ sequential words per time-step independently to others. Consequently, SAT can balance autoregressive NMT ($K=1$) and non-autoregressive NMT ($K = T$) by adjusting the value of $K$. \citet{akoury2019syntactically} moved a further step to propose a syntactically supervised Transformer (SynST), which first autoregressively predicts a chunked parse tree and then generates all words in one shot conditioned on the predicted parse.

A critical issue of NAT is that NAT copies the source words as the input of the decoder while ignores the difference between the source and target semantics. To address this problem, \citet{guo2019non} proposed to use a phrase table to covert source words to target words. They adopt a maximum match algorithm to greedily segment the source sentence into several phrases and then map these source phrases into target phrases by retrieving a pre-defined phrase table. Thanks to the enhanced decoder input, translation quality is significantly improved.

Inspired by the mask-predict task proposed by \citet{devlin2019bert}, \citet{ghazvininejad2019mask} introduced a conditioned masked language model (CMLM) to generate translation by iterative refinement. CMLM trains the conditioned language model using a mask-predict manner and produces target sentences by iterative decoding during inference. Specifically, in the training phase, CMLM first randomly masks the words in the target sentence and then predicts these masked words. In the inference, CMLM generates the entire target sentence in a preset number of decoding iteration $N$. At iteration $n \in [1, N]$, the decoder input is the entire target sentence with $T-\frac{T(N-t+1)}{N}$ words masked. The decoding process starts with a fully-masked target sentence and the words with the lowest prediction probabilities will be masked. With a proper number of decoding iteration, CMLM can effectively close the gap with fully autoregressive models and maintain the decoding efficiency.

\subsection{Alternative Training Objectives} \label{sec:objective} 
NMT trained with maximum likelihood estimation or MLE have achieved state-of-the-art results on various language pairs~\cite{junczys2016neural}. Despite the remarkable success, \citet{ranzato2015sequence} indicate two drawbacks of MLE for NMT. First, NMT models are not exposed to their errors during training, which is referred to as the \emph{exposure bias} problem. Second, MLE is defined at word-level rather than sentence-level. Due to these limitations, researchers have investigated several alternative objectives.

\citet{ranzato2016sequence} introduce Mixed Incremental Cross-Entropy Reinforce (MIXER) for sequence-level training. The MIXER algorithm borrows ideas from reinforcement learning for backpropagating gradients from non-differentiable metrics such as BLEU. \citet{D18-1397} give a study of reinforcement learning for NMT. \citet{P16-1159} proposed minimum risk training (MRT) to alleviate the problem. In MRT, the risk is defined as the expected loss with respect to the posterior distribution:
\begin{align}
\mathcal{L}(\bm{\theta}) &= \sum_{s=1}^{S} \mathbb{E}_{\mathbf{y}|\mathbf{x}^{(s)}; \bm{\theta}}\left[\Delta(\mathbf{y}, \mathbf{y}^{(s)})\right] \\
& = \sum_{s=1}^{S}\sum_{\mathbf{y} \in \mathcal{Y}(\mathbf{x}^{(s)})} P(\mathbf{y}|\mathbf{x}^{(s)}; \bm{\theta})\Delta(\mathbf{y}, \mathbf{y}^{(s)}),
\end{align}
where $\mathcal{Y}(\mathbf{x}^{(s)})$ is a set of all possible candidate translations for $\mathbf{x}^{(s)}$; $\Delta(\mathbf{y}, \mathbf{y}^{(s)})$ measures the difference between model prediction and gold-standard. \citet{P16-1159} indicate three advantages for MRT over MLE. Firstly, MRT direct optimize NMT with respect to evaluation metrics. Secondly, MRT can incorporate with arbitrary loss functions. Finally, MRT is transparent to architectures and can be applied to any end-to-end NMT systems. MRT achieves significant performance improvements than MLE training for RNNSearch. However, recent literature \cite{Choshen2020On} has also pointed out the weakness of reinforcement learning for NMT, including discussion about optimization goals and difficulty in convergence.

Efforts on improving training objectives reveal the art of translating motivation into functions and rewrite the conventional loss function with them or integrating them into it as regularizers. A collection of classical structured prediction losses are reviewed and compared in \citet{N18-1033}, including MLE, sequence-level MLE, MRT, and max-margin learning. \citet{yang-etal-2019-reducing} leveraged the idea of max-margin learning in reducing word omission errors in NMT. They artificially constructed negative examples by omitting words in target reference sentences, forcing the NMT model to assign a higher probability to a ground-truth translation and a lower probability to an erroneous translation. \citet{wieting-etal-2019-beyond} aimed at improving the semantic similarity between ground-truth references and translation outputs from NMT systems. They proposed to use a margin-based loss as an alternative reward function, encouraging NMT models to output semantically correct hypotheses even if they mismatch with the reference in the lexicon. \citet{chen2018improving} aimed at improving model capability of capturing long-range semantic structure. They proposed to explicitly model the source-target alignment with optimal transport (OT), and couple the OT loss with the MLE loss function. \citet{kumar2019von} aimed at improving model efficiency and reducing the memory footprint of NMT models. Observing that the softmax layer usually takes considerable memory usage and the longest computation time, they proposed to replace the softmax layer with a continuous embedding layer, using Von Mises-Fisher distribution to implement soft ranking as softmax layer functions. As a result, the novel probabilistic loss enables NMT models to train much faster and handle very large vocabularies.

\subsection{Using Monolingual Data and Unsupervised NMT} \label{sec:mono}
The amount of parallel data significantly affects the training of parameters as NMT is found to be data-hungry~\cite{zoph2016transfer}. Unfortunately, large-scale parallel corpora are not available for the vast majority of language pairs. In contrast, monolingual corpora are abundant and much easy to obtain. As a result, it is important to augment the training set with monolingual data.

\subsubsection{Using Monolingual Data}
As NMT is trained in an end-to-end way, it raises the difficulties in taking advantage of monolingual data. In the past few years researchers have proposed various methods to make use of the source- and target-side monolingual data in neural machine translation.

For target-side monolingual data, early attempts try to incorporate a language model trained on large-scale monolingual data into NMT. \citet{gulcehre2017integrating} proposed two ways to integrate a language model. One way is called \textit{shallow fusion}, which uses a language model during decoding to rescore the $n$-base list. Another way is called \textit{deep fusion}, which combines the decoder and language model with a controller mechanism. However, the improvements of these approaches are limited.

Another way to use target-side monolingual data is called \textit{Back-translation} (BT)~\cite{sennrich2016improving}. BT can make use of target-side monolingual data without changing the architecture of NMT. In \citet{sennrich2016improving}, they first trained a target-to-source translation model using the parallel corpus. Then, the target-side monolingual data are used to build a synthetic parallel corpus, whose source sides are generated by the target-to-source translation model. Finally, the concatenation of parallel corpus and synthetic parallel corpus is used to learn a source-to-target translation model. Although the architecture and decoding algorithm is kept unchanged, the monolingual data is fully utilized to improve the translation quality.  The authors attributed the effectiveness of using monolingual data to domain adaptation effects, reductions of overfitting, and improved fluency.
BT has shown to be the most simple and effective method to leverage target-side monolingual data~\cite{sennrich2016improving,poncelas1804investigating}. It is especially useful when only a small number of parallel data is available~\cite{karakanta2018neural}. \citet{imamura2018enhancement} found that the diversities of source sentences affect the performance of BT. In the meantime, \citet{edunov2018understanding} analyzed BT extensively and showed that noised-BT, which builds a synthetic corpus by sampled source sentences or noised output of beam-search, leads to higher accuracy. \citet{caswell2019tagged} investigated the role of noise in noised-BT. They revealed that the noises work in a way of making the model be able to distinguish the synthetic data and genuine data. The model can further take advantages of helpful signal and ignore harmful signal. As a result, they proposed a simple method called tagged-BT, which appends a preceding tag (e.g., \texttt{<BT>}) to every synthetic source sentence. \citet{wang2019improving} proposed to consider uncertainty-based confidence to help NMT models distinguish synthetic data from authentic data.

Besides target-side monolingual data, source-side monolingual data are also important resources to improve the translation quality of semi-supervised neural machine translation. \citet{zhang2016exploiting} explored two ways to leverage source-side monolingual data. The former one is knowledge distillation (also called self-training), which utilizes the source-to-target translation model to build a synthetic parallel corpus. The latter is multi-task learning that simultaneously learns translation and source sentence reordering tasks.

There are many works to make use of both source- and target-side monolingual data. \citet{hoang2018iterative} found that the translation quality of the target-to-source model in BT matters and then proposed iterative back-translation, making the source-to-target and target-to-source to enhance each other iteratively. \citet{cheng2019semi} presented an approach to train a bidirectional neural machine translation model, which introduced autoencoders on the monolingual corpora with source-to-target and target-to-source translation models as encoders and decoders by appending a reconstruction term to the training objective. \citet{he2016dual} proposed a dual-learning mechanism, which utilized reinforcement learning to make the source-to-target and target-to-source model to teach each other with the help of source- and target-side language models. \citet{zheng2019mirror} proposed a mirror-generative NMT model to integrate source-to-target and target-to-source NMT models and both-side language models, which can learn from monolingual data naturally.

\begin{figure*}[!ht]
  \centering
    \begin{subfigure}[b]{0.36\textwidth}
        \centering
        \includegraphics[width=\textwidth]{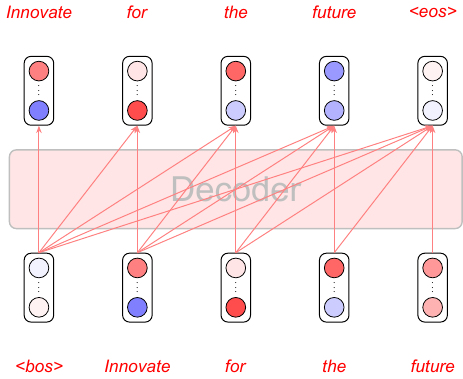}
        \caption{Unidirectional language model pre-training.}
    \end{subfigure}
    \begin{subfigure}[b]{0.36\textwidth}
        \centering
        \includegraphics[width=\textwidth]{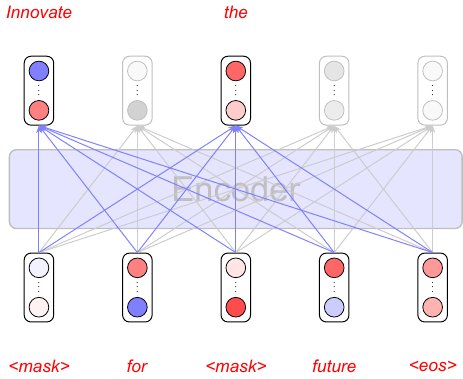}
        \caption{Bi-directional language model pre-training.}
    \end{subfigure}
    \\
    \begin{subfigure}[b]{0.663\textwidth}
    \centering
    \includegraphics[width=\textwidth]{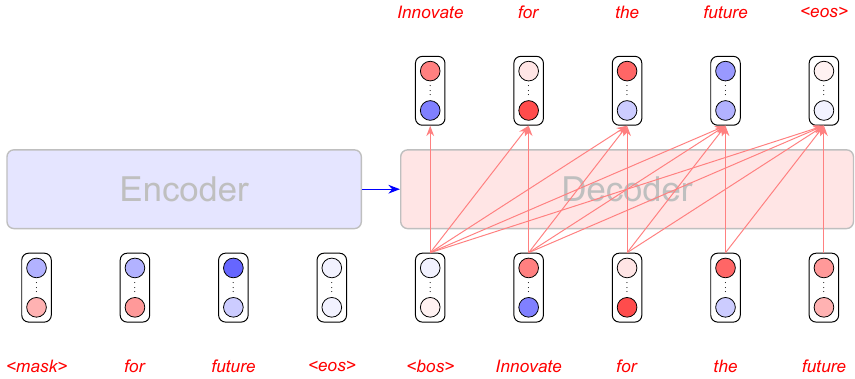}
    \caption{Sequence-to-sequence model pre-training.}
    \end{subfigure}
    \caption{Three commonly used ways for pre-training.}
  \label{fig:pre-training}
\end{figure*}

Pre-training is an alternative way to utilize monolingual data for NMT, which is shown to be beneficial by further combining with back-translation in the supervised and unsupervised NMT scenario~\cite{song2019mass,liu2020multilingual}. Recently, pre-training has attracted tremendous attention because of its effectiveness on low-resource language understanding and language generation tasks~\cite{peters2018deep, radford2019language,devlin2019bert}. Researchers found that models trained on large-scale monolingual data can learn linguistics knowledge~\cite{clark2019does}. These knowledge can be transferred into downstream tasks by initializing the task-oriented models with the pre-trained weights. Language modeling is a commonly used pre-training method. The drawback of standard language modeling is that it is unidirectional, which may be sub-optimal as a pre-training technique. \citet{devlin2019bert} proposed a masked pre-training language model (MLM) objective, which allows the model to make full use of context at the price of losing the ability to generate sequences. Combining language modeling and masking with sequence-to-sequence models, however, do not suffer from these limitations~\cite{song2019mass,lewis2019bart,liu2020multilingual}.

 \citet{edunov2019pre} fed the output representations of ELMO~\cite{peters2018deep} to the encoder of NMT. \citet{zhu2020incorporating} proposed to fuse extracted representations into each layer of encoder and decoder through attention mechanism. \citet{song2019mass} proposed to pre-train a sequence-to-sequence model first, and then finetune the pre-trained model on translation task directly. BART~\cite{lewis2019bart} took various noising method to pre-train a denoising sequence-to-sequence model and then finetune the model with an additional encoder that replaces the word embeddings of the pre-trained encoder. \citet{liu-etal-2019-robust} proposed mBART which is trained by applying BART to large-scale monolingual data across many languages.

\subsubsection{Unsupervised NMT}
Due to insufficient parallel corpus, it is not feasible to use supervised methods to train an NMT model on many language pairs. Unsupervised neural machine translation aims to obtain a translation model without using parallel data. Apparently, unsupervised machine translation is much more difficult than the supervised and semi-supervised settings.

Unsupervised neural machine translation is composed of three parts. First, by the virtue of recent advances on unsupervised cross-lingual embeddings~\cite{zhang2017adversarial,artetxe2017learning} and word-by-word translation systems~\cite{conneau2017word}, the unsupervised translation models can be initialized by weak translation models with fundamental cross-lingual information. Second, denoising autoencoders~\cite{vincent2008extracting} are used to embed the sentences into dense latent representations.  The sentences of different languages are assumed to be embedded into the same latent space so that the latent representations of source sentences can be decoded into the target language. Third, iterative back-translation is used to strengthen the source-to-target and target-to-source translation models.  \citet{lample2017unsupervised} and \citet{artetxe2017unsupervised} first successfully built an unsupervised NMT system as described above. Specifically, \citet{lample2017unsupervised} utilizes a discriminator to force the encoder to embed sentences of each language to the same latent space.

While \citet{lample2017unsupervised} used a shared encoder and a shared decoder, \citet{artetxe2017unsupervised} adopt a shared encoder but two separate decoder approach. \citet{yang2018unsupervised} conjectured that sharing of the encoder and decoder between two languages may lose their language characteristics. Therefore they proposed leveraging two separate encoders with some shared layers and using two different GANs to restrict the latent representations. \citet{artetxe2018unsupervised} and \citet{lample2018phrase} found that an unsupervised statistical machine translation system with iterative back-translation can easily outperform the unsupervised NMT counterpart. \citet{lample2018phrase} summarized that initialization, language modeling, and iterative back-translation are three principles in fully unsupervised MT and they further found that combining unsupervised SMT and unsupervised NMT can reach better performances.

\citet{ren2019unsupervised} suggested that the noises and errors existed in pseudo-data can be accumulated and hinder the improvements during iterative back-translations. Therefore, they proposed to use SMT which is less sensitive to noises as posterior regularizations to unsupervised NMT. As the unsupervised NMT is usually initialized by unsupervised bilingual word embeddings (UBWE), \citet{sun2019unsupervised} proposed to utilize UBWE agreement to enhance unsupervised NMT. \citet{wu2019extract} considered that pseudo sentences predicted by weak unsupervised MT systems are usually of low quality. To alleviate this issue, they proposed an extract-edit approach, which is an alternative to back-translation. First, they extracted the most relevant target sentences from target monolingual data given the source sentence. Then, extracted target sentences were edited to be aligned with the source sentences. This method makes it possible to use real sentence pairs to train the unsupervised NMT system. \citet{ren2020retrieve} also proposed a similar retrieve-and-rewrite method to initialize an unsupervised SMT system. \citet{artetxe2019effective} improved unsupervised SMT by exploiting subword information, developing a theoretically well-founded unsupervised tuning method, and incorporating a joint refinement procedure. Finally, they utilized the improved unsupervised SMT to initialize NMT model and get state-of-the-art results. As a unique method to utilize monolingual data, cross-lingual pre-trained models are used by \citet{lample2019cross} to initialize unsupervised MT systems.

\subsection{Open Vocabulary} \label{sec:vocab}
NMT typically operates with a fixed vocabulary. Due to practical reasons such as computational concerns and memory constraints, the vocabulary size of NMT models often ranges from 30k to 50k. For word-level NMT, the limited size of vocabulary results in a large number of unknown words. Therefore, word-level NMT is unable to translate these words and performs poorly in open-vocabulary settings~\cite{sutskever2014seq2seq,bahdanau2015nmt}.

Although word-level NMT is unable to translate out-of-vocabulary words, character-level NMT do not have this problem. By splitting words into characters, the vocabulary size is much smaller and every rare word can be represented.
\citet{chung2016character} found that the NMT model with subword-level encoder and character-level decoder can also work well.
\citet{lee2017fully} introduced a fully character-level NMT with convolutional network and found that character-to-character NMT is suitable in many-to-one multilingual setting.
\citet{luong2016achieving} built hybrid systems that translate mostly at the word level and consult the character components for rare words.
\citet{passban2018improving} proposed an extension to the model of \citet{chung2016character}, which works at the character level and boosts the decoder with target-side morphological information.
\citet{chen2018combining} proposed an NMT model at different levels of granularity with a multi-level attention.
\citet{gao2020character} found that self-attention performs very well on character-level translation.

Character-level NMT also has its imperfection, splitting words into characters results in longer sequences in which each symbol contains less information, creating both modeling and computational challenges \cite{cherry2018revisiting}. Other than word-level and character-level methods, subword-level method is another choice to model input and output sentences. \citet{sennrich2016bpe} first adapted byte-pair-encoding (BPE) to word segmentation task, which is a simple but effective method. BPE making the NMT model capable of open-vocabulary translation by encoding rare and unknown words as sequences of subword units. This method reaches a compromise between vocabulary size and sequence length with stabilized better performance over word- and character-level methods. Moreover, it is an unsupervised method with few hyper-parameters, making it the most commonly used method for word segmentation for neural machine translation and text generation. \citet{kudo2018subword} presented a simple regularization method, namely subword regularization, to improve the robustness of subword-level NMT. \citet{provilkov2019bpe} introduced BPE-dropout to regularize the subword segmentation algorithm BPE, which is more compatible with conventional BPE than the method proposed by \citet{kudo2018subword}. \citet{wang2020neural} investigated byte-level subwords, specifically byte-level BPE (BBPE), which is more efficient than using pure bytes only.

\subsection{Prior Knowledge Integration} \label{sec:knowledge}
As NMT modeling the entire translation process with a neural network, it is hard to integrate knowledge into NMT. On the one hand, existing linguistic knowledge such as dictionaries is potentially useful for NMT. On the other hand, NMT often leads to over-translation and under-translation~\cite{tu2016modeling}, which raises the need for adding prior knowledge to NMT.


One line of studies focus on inducing lexical knowledge into NMT models. \citet{Zhang:2017:ACL} proposed a general framework that can integrate prior knowledge into NMT models through posterior regularization and found that bilingual dictionary is useful to improve NMT models.
\citet{Morishita:2018:COLING} found that feeding hierarchical subword units to different modules of NMT models can also improve the translation quality.
\citet{Liu:2019:ACL} proposed a novel shared-private word embedding to capture the relationship of different words for NMT models.
\citet{Chen:2020:ACL} distinguished content words and functional words depending on the term frequency inverse document frequency (i.e., {\em TF-IDF}) and then added an additional encoder and an additional loss for content words. \citet{Weller:2020:ACL} studied strategies to model word formation in NMT to explicitly model fusional morphology.

Modeling the source-side syntactic structure has also drawn a lot of attention.
\citet{Eriguchi:2016:ACL} extended NMT models to an end-to-end syntactic model, where the decoder is softly aligned with phrases at the source side when generating a target word.
\citet{Sennrich:2016:WMT} explored external linguistic information such as lemmas, morphological features, POS tags and dependency labels to improve translation quality.
\citet{Hao:2019:EMNLP} presented a multi-granularity self-attention mechanism to model phrases which are extracted by syntactic trees.
\citet{Bugliarello:2020:ACL} proposed the Parent-Scaled Self-Attention to incorporate dependency tree to capture the syntactic knowledge of the source sentence. There are also some works that use multi-task training to learn source-side syntactic knowledge, in which the encoder of a NMT model is trained to perform POS tagging or syntactic parsing~\citep{Eriguchi:2017:ACL,Baniata:2018:AS}.

Another line of studies directly model the target-side syntactic structures~\citep{Gu:2018:EMNLP,Wang:2018:EMNLP,Wu:2017:ACL,Aharoni:2017:ACL,Bastings:2017:EMNLP,Li:2018:NAACL,Yang:2019:EMNLP,Yang:2020:ACL}. \citet{Aharoni:2017:ACL} trained a end-to-end model to directly translate source sentences into constituency trees. Similar approaches are proposed to use two neural models to generate the target sentence and its corresponding tree structure~\citep{Wang:2018:EMNLP,Wu:2017:ACL}. \citet{Gu:2018:EMNLP} proposed to use a single model to perform translation and parsing at the same time. \citet{Yang:2019:EMNLP} introduced a latent variable model to capture the co-dependence between syntax and semantics.
\citet{Yang:2020:ACL} trained a neural model to predict the soft template of the target sentence conditioning only on the source sentence and then incorporated the predicted template into the NMT model via a separate template encoder.

\subsection{Interpretability and Robustness}\label{sec:interpret}
Despite the remarkable progress, it is hard to interpret the internal workings of NMT models. All internal information in NMT is represented as high-dimensional real-valued vectors or matrices. Therefore, it is challenging to associate these hidden states with language structures. The lack of interpretability has made it very difficult for researchers to understand the translation process of NMT models.

In addition to interpretability, the lack of robustness is a severe challenge for NMT systems as well. With small perturbations in source inputs (also referred to as \emph{adversarial examples}), the translations of NMT models may lead to significant erroneous changes~\cite{belinkov2017synthetic,cheng-etal-2019-robust}. The lack of robustness of NMT limits its application on tasks that require robust performance on noisy inputs. Therefore, improving the robustness of NMT has gained increasing attention in the NMT community.

\subsubsection{Interpretability}
Efforts have been devoted to improving the interpretability of NMT systems in recent works. \citet{ding2017visualizing} proposed to visualize the internal workings of the RNNSearch~\cite{bahdanau2015nmt} architecture. With layer-wise relevance propagation~\cite{bach2015pixel}, they computed and visualized the contribution of each contextual word to arbitrary hidden states in RNNSearch. \citet{bau2018identifying} share similar motivations with \citet{ding2017visualizing}. Their basic assumption is that the same neuron in different NMT models captures similar syntactic and semantic information. They proposed to use several types of correlation coefficients to measure the importance of each neuron. As a result, by identifying important neurons and controlling their activation, the translation process of NMT systems can be controlled. \citet{strobelt2019s} also put effort into visualizing the working process of RNNSearch. The highlights of their work lie in the utilization of training data. When an NMT system decodes some words, their visualization system provides the most relevant training corpora by using the nearest neighbor search. In case of translation errors, the system can locate the erroneous outputs directly in the training set by showing its origin cause. As a result, this function provides better assistants and makes it easy for developers to adjust the model and the data.

With the tremendous success of the Transformer architecture~\cite{vaswani2017attention}, the NMT community have shown increasing interest in understanding and interpreting Transformer. \citet{he-etal-2019-towards} generalized the idea of layer-wise relevance to word importance by attributing the NMT output to every input word through a gradient-based method. The calculated word importance illustrates the influence of each source words, which also serves as an implication of under-translation errors. \citet{W18-5431} analyzed the internal representations of Transformer encoder. Utilizing the attention weights in each layer, they extract relation among each word in the source sentence. They designed four types of probing tasks to analyze the syntactic and semantic information encoded by each layer representation and test their transferability. \citet{voita-etal-2019-bottom} also proposed to analyze the bottom-up evolution of representations in Transformer with canonical correlation analysis (CCA). By estimating mutual information, they studied how information flows in Transformer.  \citet{W18-5420} proposed an operation sequence model to interpret NMT. Based on the translation outputted by the Transformer system, they proposed explicit modeling of the word reordering process and provided explicit word alignment between the reordered target-side sentence and the source sentence. As a result, one can track the reordering process of each word's information as they are explicitly aligned with the source side. Recent work~\citep{Yun2020Are} also provided a theoretical understanding of Transformer by proving that Transformer networks are universal approximators of sequence-to-sequence functions.

\subsubsection{Robustness}
\citet{belinkov2017synthetic} first investigated the robustness of NMT. They pointed out that both synthetic and natural noise can severely harm the performance of NMT models. They experimented with four types of synthetic noise and leveraged structure-invariant representation and adversarial training to improve the robustness of NMT. Similarly, \citet{zhao2017generating} proposed to map the input sentence to a latent space with generative adversarial networks (GAN) and search for adversarial examples in that space. Their approach can produce semantically and syntactically coherent sentences that have negative impacts on the performance of NMT models.

\citet{P18-1079} proposed semantic-preserving adversarial rules to explicitly induce adversarial examples. This approach provides a better guarantee for the adversarial examples to satisfy semantically equivalence property. \citet{P18-1163} proposed two types of approaches to generating adversarial examples by perturbing the source sentence or the internal representation of the encoder. By integrating the effect of adversarial examples into the loss function, the robustness of neural machine translation is improved by adversarial training.

\citet{C18-1055} proposed a character-level white-box attack for character-level NMT. They proposed to model the operations of character insertion, deletion, and swapping with vector computations so that the generation of adversarial examples can be formulated with differentiable string-edit operations. \citet{liu-etal-2019-robust} proposed to jointly utilize textual and phonetic embedding in NMT to improve robustness. They found that to train a more robust model, more weights should be put on the phonetic rather than textual information. \citet{cheng-etal-2019-robust} proposed doubly adversarial inputs to improve the robustness of NMT. Concretely, they proposed to both attack the translation model with adversarial source examples and defend the translation model with adversarial target inputs for model robustness. \citet{zou-etal-2020-reinforced} utilized reinforcement learning to generate adversarial examples, producing stable attacks with semantic-preserving adversarial examples. \citet{cheng-etal-2020-advaug} proposed a novel adversarial augmentation method that minimizes the vicinal risk over virtual sentences sampled from a smoothly interpolated embedding space around the observed training sentence pairs. The adversarial data augmentation method substantially outperforms other data augmentation methods and achieves significant improvements in translation quality and robustness. For the better exploration of robust NMT, \citet{D18-1050} proposed an MTNT dataset, source sentences of which are collected from Reddit discussion, and contain several types of noise. Target referenced translations for each source sentence, in contrast, are clear from noise. Experiments showed that current NMT models perform badly on the MTNT dataset. As a result, this dataset can serve as a testbed for NMT robustness analysis.

\section{Resources}\label{sec:resources}

\subsection{Parallel Data}

\begin{table*}[t]
\centering
\begin{tabular}{c|l|l}
\toprule
\textbf{Workshop} & \textbf{Domain} & \textbf{Language Pair} \\ \hline
\multirow{3}{*}{WMT20} & News & zh-en, cz-en, fr-de, de-en, iu-en, km-en, ja-en, ps-en, pl-en, ru-en, ta-en \\
& Biomedical & en-eu, en-zh, en-fr, en-de, en-it, en-pt, en-ru, en-es\\
& Chat & en-de \\
\hline
\multirow{3}{*}{IWSLT20} & TED Talks & en-de \\
& e-Commerce & zh-en, en-ru \\
& Open Domain & zh-ja \\
\hline
\multirow{5}{*}{WAT20} & Scientific Paper & en-ja, zh-ja\\
& Business Scene Dialogue & en-ja \\
& Patent & zh-ja, ko-ja, en-ja \\
& News & ja-en, ja-ru \\
& IT and Wikinews & hi-en, th-en, ms-en, id-en \\
\bottomrule
\end{tabular}
\caption{Domain and language pairs provided by WMT20, IWSLT20, WAT20.}
\label{tab:resources}
\end{table*}

\begin{table*}[]
\centering
\begin{tabular}{l|c|c|c|c|c|c|c|c|c}
\toprule
\textbf{Source} & \textbf{Fr-En} & \textbf{Es-En} & \textbf{De-En} & \textbf{Pt-En} & \textbf{Ru-En} & \textbf{Ar-En} & \textbf{Zh-En} & \textbf{Ja-En} & \textbf{Hi-En} \\ \hline
OPUS~\cite{tiedemann2016opus} & 200.6M & 172.0M & 93.3M & 77.7M & 75.5M & 69.2M & 31.2M & 6.2M & 1.7M \\\bottomrule
\end{tabular}
\caption{Number of sentences that available at OPUS for major languages to English.}
\label{tab:opus-resources}
\end{table*}

Bilingual parallel corpora are the most important resources for NMT. There are several publicly available corpora, such as the datasets provided by WMT\footnote{\url{http://www.statmt.org/wmt20/index.html}}, IWSLT\footnote{\url{http://iwslt.org/doku.php}}, and WAT~\footnote{\url{http://lotus.kuee.kyoto-u.ac.jp/WAT/WAT2020/index.html}}. Table~\ref{tab:resources} lists the available domains and language pairs in these workshops.

Besides the aforementioned machine translation workshops, we also recommend OPUS~\footnote{\url{http://opus.nlpl.eu}} to search resources for training NMT models, which gathers parallel data for a large number of language pairs. We list the number of sentence pairs that are available for major languages to Enligsh in Table~\ref{tab:opus-resources}. OPUS also provides the OPUS-100 corpus for multilingual machine translation research~\cite{zhang2020improving}, which is an English-centric multilingual corpus covering over 100 languages.

\subsection{Monolingual Data}

Monolingual data are also valuable resources for NMT. The Common Crawl Foundation~\footnote{\url{https://commoncrawl.org/}} provides open access to high quality crawled data for over 40 languages. The \textsc{CCNet} toolkit~\footnote{\url{https://github.com/facebookresearch/cc_net}}~\cite{wenzek2020ccnet} can be used to download and clean Common Crawl texts. Wikipedia provides database dump~\footnote{\url{https://dumps.wikimedia.org}} that can be used to extract monolingual data, which can be download using \textsc{WikiExtractor}~\footnote{\url{https://github.com/attardi/wikiextractor}}. WMT 2020 also provides several monolingual training data, which consists of data collected from NewsCrawl, NewsDicussions, Europarl, NewsCommentary, CommonCrawl, and WikiDumps.

\section{Tools} \label{sec:tools}
With the rapid advances of deep learning, many open-source deep learning frameworks have emerged, with TensorFlow~\cite{abadi2016tensorflow} and PyTorch~\cite{paszke2019pytorch} as representative examples. At the same time, we have also witnessed the rapid development of open-source NMT toolkits, which significantly boosted the research progress of NMT. In this section, we will give a summarization of popular open-source NMT toolkits. Besides, we also introduce tools that are useful for evaluation, analysis, and data pre-processing.

\subsection{Open-source NMT Toolkits}
\begin{table*}[t!]
\centering
\begin{tabular}{c|c|c|c}
\toprule
{\bf Name} & {\bf Language} & {\bf Framework} & {\bf Status} \\\hline
\href{https://github.com/tensorflow/tensor2tensor}{\textsc{Tensor2Tensor}} & Python & TensorFlow & Deprecated \\
\href{https://github.com/pytorch/fairseq}{\textsc{FairSeq}} & Python & PyTorch & Active \\
\href{https://github.com/tensorflow/nmt}{\textsc{Nmt}} & Python & TensorFlow & Deprecated \\
\href{https://github.com/OpenNMT}{\textsc{OpenNMT}} & Python/C++ & PyTorch/TensorFlow & Active \\
\href{https://github.com/awslabs/sockeye}{\textsc{Sockeye}} & Python & MXNet & Active \\
\href{https://github.com/EdinburghNLP/nematus}{\textsc{Nematus}} & Python & Tensorflow & Active \\
\href{https://github.com/marian-nmt/marian}{\textsc{Marian}} & C++ & - & Active \\
\href{https://github.com/THUNLP-MT/THUMT}{\textsc{THUMT}} & Python & PyTorch/TensorFlow & Active \\
\href{https://github.com/lvapeab/nmt-keras}{\textsc{NMT-Keras}} & Python & Keras & Active \\
\href{https://github.com/ufal/neuralmonkey}{\textsc{Neural Monkey}} & Python & TensorFlow & Active \\
\bottomrule
\end{tabular}
\caption{Popular Open-source NMT toolkits on GitHub, the ordering is determined by the number of stars as the date of December 2020.}
\label{tab:tools}
\end{table*}

We summarize some popular open-source NMT toolkits on GitHub in Table~\ref{tab:tools}.
The users can get the source codes of these toolkits directly from GitHub. We shall give a brief description of these projects.

\textbf{Tensor2Tensor.}~\textsc{Tensor2Tensor}~\cite{vaswani-2018-tensor2tensor}  is a library of deep learning models and datasets based on TensorFlow~\cite{abadi2016tensorflow}. The library was mainly developed by the Google Brain team. \textsc{Tensor2Tensor} provides implementation of several NMT architectures (e.g., Transformer) for the translation task. The users can run \textsc{Tensor2Tensor} easily on CPU, GPU, and TPU, either locally or on Cloud.

\textbf{FairSeq.}~\textsc{FairSeq}~\cite{ott2019fairseq} is a sequence modeling toolkit developed by Facebook AI Research. The toolkit is based on Pytorch~\cite{paszke2019pytorch} and allows the users to train custom models for the translation task. \textsc{FairSeq} implements traditional RNN-based models and Transformer models. Besides, it also includes CNN-based translation models (e.g., LightConv and DynamicConv).

\textbf{Nmt.}~\textsc{Nmt}~\cite{luong17} is a toolkit developed by Google Research. The toolkit implements the GNMT architecture~\cite{wu2016gnmt}. Besides, the NMT project also provides a nice tutorial for building a competitive NMT model from scratch. The codebase of NMT is high-quality and lightweight, which is friendly for users to add customized models.

\textbf{OpenNMT.}~\textsc{OpenNMT} is an open-source NMT toolkit developed by the collaboration of Harvard University and SYSTRAN. The toolkit currently maintains two implementations: \textsc{OpenNMT-py} and \textsc{OpenNMT-tf}. \textsc{OpenNMT} is proven to be research-friendly and production-ready. The OpenNMT project also provides \textsc{CTranslate2} as a fast inference engine that supports both CPU and GPU.

\textbf{Sockeye.}~\textsc{Sockeye}~\cite{hieber2017sockeye} is a versatile sequence-to-sequence toolkit that is based on MXNet~\cite{chen2016mxnet}. \textsc{Sockeye} is maintained by Amazon and powers machine translation services such as Amazon Translate. The toolkit features state-of-the-art machine translation models and fast CPU inference, which is useful for both research and production.

\textbf{Nematus.}~\textsc{Nematus} is an NMT toolkit developed by the NLP Group at the University of Edinburgh. The toolkit is based on TensorFlow and supports RNN-based NMT architectures as well as the \textsc{Transformer} architecture. In addition to the toolkits, \textsc{Nematus} also released high-performing NMT models covering 13 translation directions.

\textbf{Marian.}~\textsc{Marian}~\cite{junczys-dowmunt-etal-2018-marian} is an efficient and self-contained NMT framework currently being developed by the Microsoft Translator team. The framework is written entirely in C++ with minimal dependencies. Marian is widely deployed by many companies and organizations. For example, Microsoft Translator currently adopts Marian as its neural machine translation engine.

\textbf{THUMT.}~\textsc{THUMT}~\cite{tan-etal-2020-thumt} is an open-source toolkit for neural machine translation developed by the NLP Group at Tsinghua University. The toolkit includes TensorFlow, and Pytorch implementations. It supports vanilla RNN-based and Transformer models and is easy for users to build new models. Furthermore, \textsc{THUMT} provides visualization analysis using layer-wise relevance propagation~\cite{ding2017visualizing}.

\textbf{NMT-Keras.}~\textsc{NMT-Keras}~\cite{nmt-keras18} is a flexible toolkit for neural machine translation developed by the Pattern Recognition and Human Language Technology Research Center at Polytechnic University of Valencia. The toolkit is based on Keras which uses Theano or TensorFlow as the backend. \textsc{NMT-Keras} emphasizes the development of advanced applications for NMT systems, such as interactive NMT and online learning. It also has been extended to other tasks including image and video captioning, sentence classification, and visual question answering.

\textbf{Neural Monkey}~\textsc{Neural Monkey} is an open-source neural machine translation and general sequence-to-sequence learning system. The toolkit is built on the TensorFlow library and provides a high-level API tailored for fast prototyping of complex architectures.


\subsection{Tools for Evaluation and Analysis}

Manual evaluation of MT outputs is not only expensive but also impractical to scaling for large language pairs. On the contrary, automatic MT evaluation is inexpensive and language-independent, with BLEU~\cite{papineni2002bleu} as the representative automatic evaluation metric. Besides evaluation, there is also a need for analyzing MT outputs. We recommend the following tools for evaluating and analyzing MT output.

 \textbf{\textsc{Sacre}BLEU.} \textsc{SacreBLEU}~\footnote{\url{https://github.com/mjpost/sacrebleu}}~\cite{post2018call} is a toolkit to compute shareable, comparable, and reproducible BLEU scores. \textsc{SacreBLEU} computes BLEU scores on detokenized outputs, using WMT standard tokenization. As a result, the scores are not affected by different processing tools. Besides, it can produce a short version string that facilitates cross-paper comparisons.

\textbf{\textsc{Compare-mt}.} \textsc{Compare-mt}\footnote{\url{https://github.com/neulab/compare-mt}}~\cite{Neubig:2019:NAACL} is a program to compare the outputs of multiple systems for language generation. In order to provide high-level analysis of outputs, it enables analysis of accuracy of generation of particular types of words, bucketed histograms of sentence accuracies or counts based on salient characteristics, and so on.

\textbf{\textsc{MT-ComparEval}.} \textsc{MT-ComparEval}\footnote{\url{https://github.com/ondrejklejch/MT-ComparEval}} is also a tool for comparison and evaluation of machine translations. It allows users to compare translations according to automatic metrics or quality comparison from the aspects of n-grams.

\subsection{Other Tools}
Asides from the above mentioned tools, we found the following toolkits are very useful for NMT research and deployment.

\textbf{\textsc{Moses}.} \textsc{Moses}\footnote{\url{https://github.com/moses-smt/mosesdecoder}}\cite{koehn2007moses} is a self-contained statistical machine translation toolkit. Besides SMT-related components, \textsc{Moses} provides a large number of tools to clean and pre-process texts, which are also useful for training NMT models. \textsc{Moses} also contains several easy-to-use scripts to analyze and evaluate MT outputs.

\textbf{\textsc{Subword-NMT}.}
\textsc{Subword-NMT}\footnote{\url{https://github.com/rsennrich/subword-nmt}} is an open-source toolkit for unsupervised word segmentation for neural machine translation and text generation. It adopts the Byte-Pair Encoding (BPE) algorithm proposed by~\cite{sennrich2016bpe} and BPE dropout proposed by~\cite{provilkov2019bpe}. It is the most commonly used toolkit to alleviate the out-of-vocabulary problem in NMT.

\textbf{\textsc{SentencePiece}.} \textsc{SentencePiece}\footnote{\url{https://github.com/google/sentencepiece}} is a powerful unsupervised text segmentation toolkit. \textsc{SentencePiece} is written in C++ and provides APIs for other languages such as Python. \textsc{SentencePiece} implements the BPE algorithm~\cite{sennrich2016bpe} and unigram language model~\cite{kudo2018subword}. Unlike \textsc{Subword-NMT}, \textsc{SentencePiece} can learn to segment raw texts without additional pre-processing. As a result, \textsc{SentencePiece} is a suitable choice to segment multilingual texts.

\section{Conclusion}~\label{sec:conclusion}
Neural machine translation has become the dominant approach to machine translation in both research and practice. This article reviewed the widely used methods in NMT, including modeling, decoding, data augmentation, interpretation, as well as evaluation. We then summarize the resources and tools that are useful for NMT research.

Despite the great success achieved by NMT, there are still many problems to be explored. We list some important and challenging problems for NMT as follows:
\begin{itemize}
\item \textbf{Understanding NMT.} Although there are many attempts to analyze and interpret NMT, our understandings about NMT are still limited. Understanding how and why NMT produces its translation result is important to figure out the bottleneck and weakness of NMT models.
\item \textbf{Designing better architectures}. Designing a new architecture that better than Transformer is beneficial for both NMT research and production. Furthermore, designing a new architecture that balances translation performance and computational complexity is also important.
\item \textbf{Making full use of monolingual data.} Monolingual data are valuable resources. Despite the remarkable progress, we believe that there is still much room for NMT to make use of abundant monolingual data.
\item \textbf{Integrating prior knowledge.} Incorporating human knowledge into NMT is also an important problem. Although there is some progress, the results are far from satisfactory. How to convert discrete and continuous representations into each other is a problem of NMT that needs further exploration.
\end{itemize}

\section*{Acknowledgements}
This work was supported by the National Key R\&D Program of China (No. 2017YFB0 202204), National Natural Science Foundation of China (No. 61925601, No. 61761166 008, No. 61772302), Beijing Academy of Artificial Intelligence, Huawei Noah's Ark Lab, and the NExT++ project supported by the National Research Foundation, Prime Ministers Office, Singapore under its IRC@Singapore Funding Initiative. We thank Kehai Chen and Xiangwen Zhang for their kindly suggestions.

\bibliography{nmt}

\end{document}